\pdfoutput=1
\documentclass[11pt]{article}

\usepackage[preprint]{acl}

\usepackage{times}
\usepackage{latexsym}
\usepackage{amsmath}
\usepackage{booktabs}

\usepackage[T1]{fontenc}
\usepackage[utf8]{inputenc}

\usepackage{microtype}

\usepackage{inconsolata}
\usepackage{caption}
\usepackage{stfloats}
\usepackage{xspace}

\usepackage{graphicx}
\usepackage{tabularx}
\usepackage{tcolorbox}
\usepackage{adjustbox}
\usepackage{bm}
\usepackage{subfigure}
\usepackage{soul}
\usepackage{color}
\usepackage{comment}
\usepackage{colortbl}

\newcommand{\hlred}{\colorlet{c}{red!20}\sethlcolor{c}\hl}
\newcommand{\hlgreen}{\colorlet{c}{green!20}\sethlcolor{c}\hl}
\newcommand{\hlyellow}{\colorlet{c}{yellow!20}\sethlcolor{c}\hl}
\newcommand{\hlblack}{\colorlet{c}{black!20}\sethlcolor{c}\hl}

\title{A Unified Agentic Framework for Evaluating Conditional Image Generation}

\newcommand{\score}{\textsc{CIGEval}\xspace}

\author{
  Jifang Wang, Xue Yang, Longyue Wang, Zhenran Xu, Yiyu Wang, \\
  \textbf{Yaowei Wang, Weihua Luo, Kaifu Zhang, Baotian Hu\thanks{Corresponding author.}, Min Zhang} \\
  Harbin Institute of Technology (Shenzhen), Shenzhen, China\\
  \texttt{23S151116@stu.hit.edu.cn, \{hubaotian,zhangmin2021\}@hit.edu.cn} \\
}

\begin{document}

\maketitle
\begin{abstract}
Conditional image generation has gained significant attention for its ability to personalize content. 
However, the field faces challenges in developing task-agnostic, reliable, and explainable evaluation metrics.
This paper introduces \textbf{\score}, a unified agentic framework for comprehensive evaluation of conditional image generation tasks. 
\score utilizes large multimodal models (LMMs) as its core, integrating a multi-functional toolbox and establishing a fine-grained evaluation framework. Additionally, we synthesize evaluation trajectories for fine-tuning,
empowering smaller LMMs to autonomously select appropriate tools and conduct nuanced analyses based on tool outputs. 
Experiments across seven prominent conditional image generation tasks demonstrate that \score (GPT-4o version) achieves a high correlation of 0.4625 with human assessments, closely matching the inter-annotator correlation of 0.47. 
Moreover, when implemented with 7B open-source LMMs using only 2.3K training trajectories, \score surpasses the previous GPT-4o-based state-of-the-art method.
Case studies on GPT-4o image generation highlight \score's capability in identifying subtle issues related to subject consistency and adherence to control guidance,
indicating its great potential for automating evaluation of image generation tasks with human-level reliability\footnote{Our code and models are publicly available at \url{https://github.com/HITsz-TMG/Agentic-CIGEval}.}.

% Conditional image generation has gained significant attention for its ability to personalize content. However, the field faces challenges in developing task-agnostic, reliable, and explainable evaluation metrics. This paper introduces CIGEval, a unified agentic framework for comprehensive evaluation of conditional image generation tasks. CIGEval utilizes large multimodal models (LMMs) as its core, integrating a multi-functional toolbox and establishing a fine-grained evaluation framework. Additionally, we synthesize evaluation trajectories for fine-tuning, empowering smaller LMMs to autonomously select appropriate tools and conduct nuanced analyses based on tool outputs. Experiments across seven prominent conditional image generation tasks demonstrate that CIGEval (GPT-4o version) achieves a high correlation of 0.4625 with human assessments, closely matching the inter-annotator correlation of 0.47. Moreover, when implemented with 7B open-source LMMs using only 2.3K training trajectories, CIGEval surpasses the previous GPT-4o-based state-of-the-art method. Case studies on GPT-4o image generation highlight CIGEval's capability in identifying subtle issues related to subject consistency and adherence to control guidance, indicating its great potential for automating evaluation of image generation tasks with human-level reliability.

\end{abstract}

\section{Introduction}

\begin{figure}[!t]
    \centering
    \includegraphics[width=\linewidth]{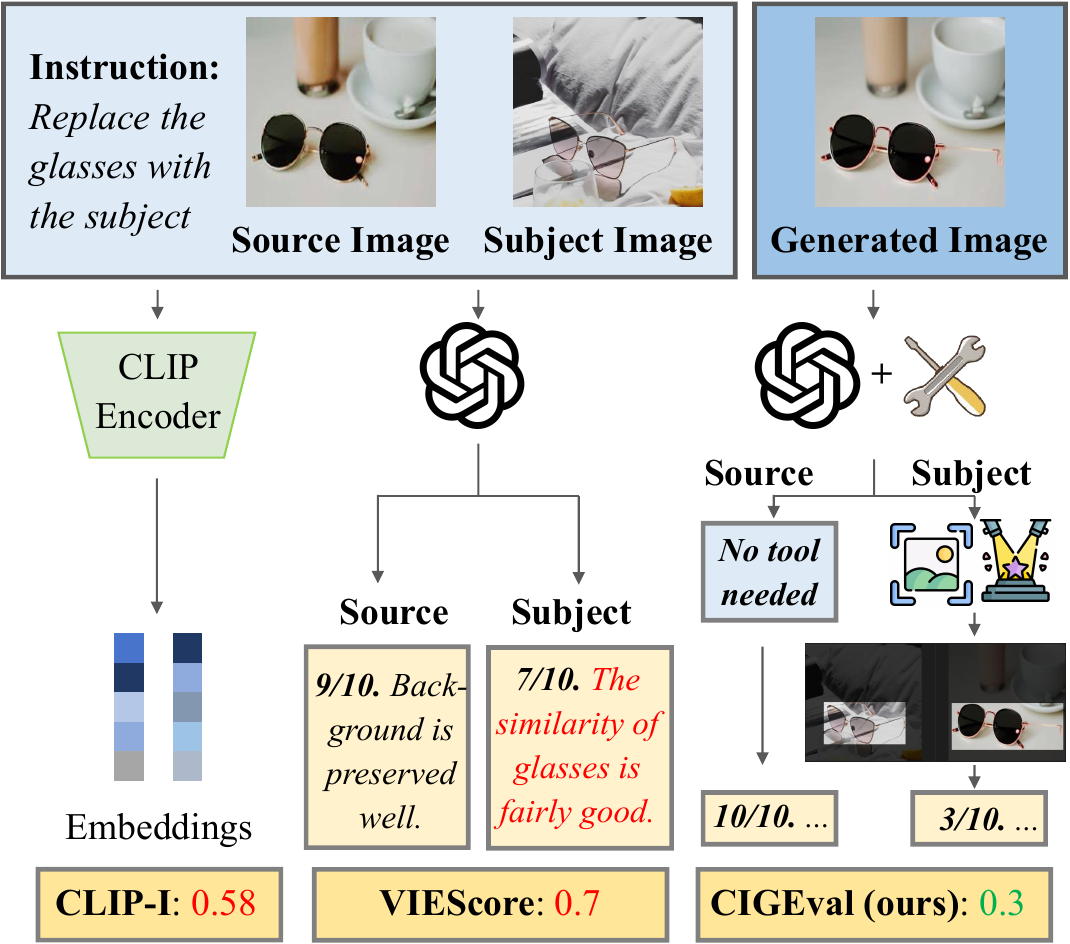}
    \caption{An example of subject-driven image editing with human-annotated low scores.
    Both traditional metrics and GPT-4o-based VIEScore assign high scores. 
     By integrating GPT-4o with tools, \score, our agentic evaluation framework, highlights the glasses object in both images, and finds their different shapes and designs, thereby reaching the correct score. ``Source'' and ``Subject'' means ``source image'' and ``subject image''.}
    \label{fig:intro}
\end{figure}

Recent advances in large-scale text-to-image (T2I) generative models have enabled the creation of images based on text prompts as well as reference images, i.e. \textit{conditional} image generation~\cite{kumari2023multi,ruiz2023dreambooth,li2023dreamedit,he2024imagineyourself}.
The field is evolving at an unprecedented pace with an increasing number of tasks and models being introduced. 
Among these, text-guided image generation is particularly popular~\cite{ramesh2022hierarchical,chen2025januspro,yuan-etal-2025-finerag}.
% which aims to ground on a text prompt to generate the corresponding image.
Expanding beyond text, a diverse set of conditions have been employed to steer the diffusion process: 
text-guided image editing~\cite{brooks2022instructpix2pix}, 
mask-guided image editing~\cite{sdinpaint}, 
subject-driven image generation and editing~\cite{chen2023subject,guo2024pulid}, multi-concept image composition~\cite{kumari2023multi} 
and control-guided image generation~\cite{qin2023unicontrol,zhang2023adding}.

Despite the growing number of generative models being developed, 
a significant challenge persists in effectively evaluating AI-synthesized images~\cite{peng2024dreambench}.
Current evaluation metrics have the following three limitations:
(1) \textbf{Task-specific}: 
Traditional metrics are narrowly focused and cannot be generalized across different tasks. 
For example, LPIPS~\cite{zhang2018perceptual} measures the perceptual similarity of a pair of images, while CLIP-Score~\cite{hessel2021clipscore} measures the text alignment of one single image.
(2) \textbf{Limited explainability}: 
Assigning a single score to a generated image without the reasoning process fails to offer a comprehensive evaluation.
Images can be assessed on multiple dimensions, such as prompt adherence and concept preservation~\cite{fu2023dreamsim}.
(3) \textbf{Lack of human alignment}: Traditional metrics like DINO~\cite{Caron2021EmergingPI} and CLIP~\cite{pmlr-v139-radford21a} often result in huge discrepancies from humans, caused by their image similarity measurement nature.
Even based on the powerful large multimodal model (LMM) GPT-4o, 
as shown in Figure~\ref{fig:intro},
the current state-of-the-art VIEScore~\cite{ku-etal-2024-viescore} 
struggles to capture subtle image nuances and 
shows low correlation with human judgment in various image editing tasks.

To address these issues,
we propose \textbf{\score},
an autonomous LMM-based agent framework for evaluating conditional image generation.
This agent framework can integrate the advanced GPT-4o model~\cite{openai2023gpt4} and open-source models (\textit{e.g.}, Qwen2.5-VL~\cite{wang2024qwen2vl}).
Our work is driven by two primary motivations: 
(1) developing autonomous evaluation agents capable of making independent decisions and judgments without human assistance;
(2) enabling relatively smaller models to efficiently perform complex evaluations.
To achieve this,
we make three key technical contributions.
First, we extend the LMM's capability to detect and emphasize subtle differences between highly similar images by curating a versatile toolbox,
in contrast to previous methods that relied solely on the perceptual capabilities of LMMs.
Second, we establish a fine-grained evaluation framework, including task decomposition, tool selection and analysis.
Third, we synthesize instruction data based on evaluation trajectories for fine-tuning the LMM,
where we first employ GPT-4o to execute the stages and then filter the trajectories that align with human evaluations.

Experiments on the well-established ImagenHub benchmark~\cite{ku2023imagenhub} show that, 
when using GPT-4o as the underlying LMM,
\score achieves the state-of-the-art performance across all 7 tasks.
It achieves an average Spearman correlation of 0.4625 with human raters,
closely matching the human-to-human correlation of 0.47.
The primary improvements are observed in tasks involving multiple conditions, 
such as control-guided image generation and multi-concept image composition, 
where previous evaluation metrics struggle.
Using only 2.3K filtered evaluation trajectories for tuning,
\score, leveraging 7B open-source LMMs, 
demonstrates performance surpassing previous GPT-4o-based state-of-the-art methods.
Further ablation study shows the importance of each tool and the robustness of our framework.
In addition, we conduct a preliminary case study on GPT-4o's image generation.
\score assigns scores closely aligned with human annotations 
and effectively detects subtle flaws in 4o-generated images, 
especially in tasks involving multiple input images and adherence to specific control signals (e.g., Canny edges, OpenPose).
These results suggest that \score has substantial promise for achieving human-level performance in assessing synthetic images.

Our main contributions are as follows:
\begin{itemize}
    \item We introduce \score, an LMM-based evaluation agent designed to assess various conditional image generation tasks. Our approach is characterized by its human-aligned, explainable, and unified evaluation method, setting it apart from previous metrics.
    \item We evaluate \score across 7 conditional image generation tasks, demonstrating that \score, based on GPT-4o, outperforms all existing baselines and achieves a high correlation with human annotators, closely mirroring the human-to-human correlation.
    \item We fine-tune open-sourced 7B LMMs and significantly improve their evaluation performance, surpassing previous GPT-4o-based state-of-the-art method.
\end{itemize}

\section{Related Work}

\subsection{Conditional Image Generation}

Diffusion models have gained wide attention in AI research for image synthesis~\cite{NEURIPS2020_4c5bcfec_diffusion,NEURIPS2021_49ad23d1_diffusion}. 
In recent years, several new models \citep{kumari2023multi,ruiz2023dreambooth, li2023dreamedit, zhang2023adding} have been developed to introduce controllable conditions in image generation. 
Prevalent tasks in this domain include text-to-image generation \cite{saharia2022photorealistic, rombach2022high, sd-xl} (known as text-guided image generation), inpainting \citep{avrahami2022blended, Lugmayr2022RePaintIU} (referred to as mask-guided image editing) and text-guided image editing \citep{brooks2022instructpix2pix, couairon2022diffedit, cyclediffusion}. 
Recent works have proposed new tasks,
such as subject-driven image generation and editing \citep{gal2022image, ruiz2023dreambooth, li2023dreamedit} to inject one specific subject into a synthesized image, 
and multi-concept image composition \citep{kumari2023multi, liu2023cones,ding2024freecustom}, which allows multiple specific subjects into the synthesized image. 
Additionally, control-guided image generation~\citep{zhang2023adding, qin2023unicontrol,guo2024pulid} allows additional conditions alongside the text prompt to guide the image synthesis. 
Our work employs LMM-based agents to assess all of these discussed tasks.

\subsection{Synthetic Image Evaluation}

A variety of metrics have been introduced to assess AI-generated images. 
For example, the CLIP score \citep{hessel2021clipscore} and BLIP score \citep{li2022blip} are commonly used to measure the alignment between the generated image and the text prompt.
Metrics like LPIPS~\cite{zhang2018perceptual} and DreamSim~\cite{fu2023dreamsim}  focus on assessing perceptual similarity. 
LLMScore~\cite{lu2023llmscore} and HEIM-benchmark \citep{lee2023holistic} assess text-to-image models on multiple fine-grained aspects, including toxicity and safety.
However, these metrics predominantly focused on text-to-image generation and remain narrow in scope. 
There is a noticeable lack of effective automatic metrics for other conditional image generation tasks, such as subject-driven image generation and image editing \citep{ruiz2023dreambooth, li2023dreamedit,peng2024dreambench}. 
Consequently, 
some research work~\citep{denton2015deep, isola2017image, meng2021sdedit, chen2023subject, sheynin2023emu} rely heavily on human evaluation.
This dependence highlights the need for more unified, interpretable, and reliable automatic evaluation methods in the field.
Our work seeks to bridge this gap by developing an autonomous agentic evaluation framework that closely aligns with human judgment.

\subsection{Large Multimodal Models as Evaluators} 

Motivated by the explorations of large language model (LLM)-based
evaluators in natural language processing~\cite{zheng2023judging,dubois2023alpacafarm,Fu2023GPTScoreEA,cheng-etal-2024-small}, 
large multimodal models (LMMs) have been utilized to evaluate responses in visual question answering~\cite{mllm_as_judge,xu-etal-2024-multiskill}.
In the realm of image evaluation,
the GPT-4 series has demonstrated promising capabilities, particularly in assessing text-image alignment \citep{Zhang2023GPT4visionAA,li2024gpt4_eval}. 
However, these models are not without limitations.
A comprehensive study on GPT-4o's vision abilities have revealed mistakes in fine-grained image evaluation tasks \citep{yang2023dawn},
such as failing to accurately distinguish differences between similar images~\cite{ku-etal-2024-viescore}.
To address these shortcomings, 
we enhance the capabilities of LMMs by integrating a versatile set of image analysis and editing tools,
and by adopting an agentic framework to improve the evaluation of AI-generated images.

\section{\score}

In this section,
we introduce \textbf{\score}, our LMM-based agentic framework designed for evaluating conditional image generation.
First, we define seven conditional image generation tasks that are the focus of our study (Sec.~\ref{sec:tasks}),
and then design a multi-functional toolbox (Sec.~\ref{sec:tools}).
Next, we introduce our fine-grained evaluation framework (Sec.~\ref{sec:framework}).
Finally, we synthesize high-quality trajectory data to fine-tune open-source LMMs (Sec.~\ref{sec:tuning}).

\subsection{Task Definition}
\label{sec:tasks}

To build a unified and explainable evaluation metric, 
we define the image evaluation problem as shown in Equation \ref{eq:VIE}.
The function $f_\text{eval}$ takes as input an instruction $I$, a synthesized image $O$, and a set of conditions $C^*$ (e.g. text prompt, subject image, background image, canny-edge, etc). 
The function $f_\text{eval}$ should produce the intermediate rationale in natural language before generating the final score according to the instruction $I$:
\begin{equation}
    f_\text{eval}(I, O, C^*) = (\text{rationale}, \text{score}) %\propto \text{Human}
    \label{eq:VIE}
\end{equation}

Following \citet{ku2023imagenhub}, we focus on seven primary conditional image generation tasks, each with different sets of conditions $C^*$:

\textbf{• Text-guided Image Generation}: $C^* = [p]$, where $p$ is a text prompt. The objective is to generate an image that aligns with the text description.

\textbf{• Mask-guided Image Editing}: $C^* = [p, I_\text{mask}, $ $I_\text{src}]$, where $I_\text{mask}$ is a binarized mask and $I_\text{src}$ is a source image. The aim is to modify $I_\text{src}$ in the masked area according to $p$.

\textbf{• Text-guided Image Editing}: $C^* = [p, I_\text{src}]$. This task is similar to Mask-guided Image Editing but does not provide a mask. The model must identify the region to edit automatically.

\textbf{• Subject-driven Image Generation}: $C^* = [p, $ $S]$, where $S$ is the image of a specific subject. The aim is to generate an image that reflects $p$ in relation to the subject $S$.

\textbf{• Subject-driven Image Editing}: $C^* = [S, p, $ $I_\text{src}]$, where $I_\text{src}$ is a source image, and $S$ is the subject reference. The goal is to replace the subject in $I_\text{src}$ with $S$.

\textbf{• Multi-concept Image Composition}: $C^* = [S_1, S_2, p, I_\text{src}]$, where $S_1$ and $S_2$ are images of two subjects. The task is to combine these to create a new image according to $p$.

\textbf{• Control-guided Image Generation}: $C^* = [I_\text{control}, p]$, where $I_\text{control}$ is a control signal, such as a depth map, canny edge, or bounding box. The aim is to generate an image that follows these low-level visual cues.

In this paper, following previous work~\cite{mañas2024texttoimage,lin2024vqascore}, we investigate the semantic consistency of generated images with the above conditions.

\subsection{Toolbox}
\label{sec:tools}

\begin{table*}[ht]
%\small
\centering
\begin{adjustbox}{max width=\linewidth}
\begin{tabular}{@{}p{70pt}|p{50pt}p{110pt}|p{144pt}p{1pt}|p{100pt}@{}}
\toprule
\textbf{Tool}  & \multicolumn{2}{c|}{\textbf{Argument}} & \multicolumn{2}{c|}{\textbf{Output}} & \textbf{Purpose}\\
\midrule
\textbf{\textsf{Grounding}} \newline \includegraphics[width=0.08\textwidth]{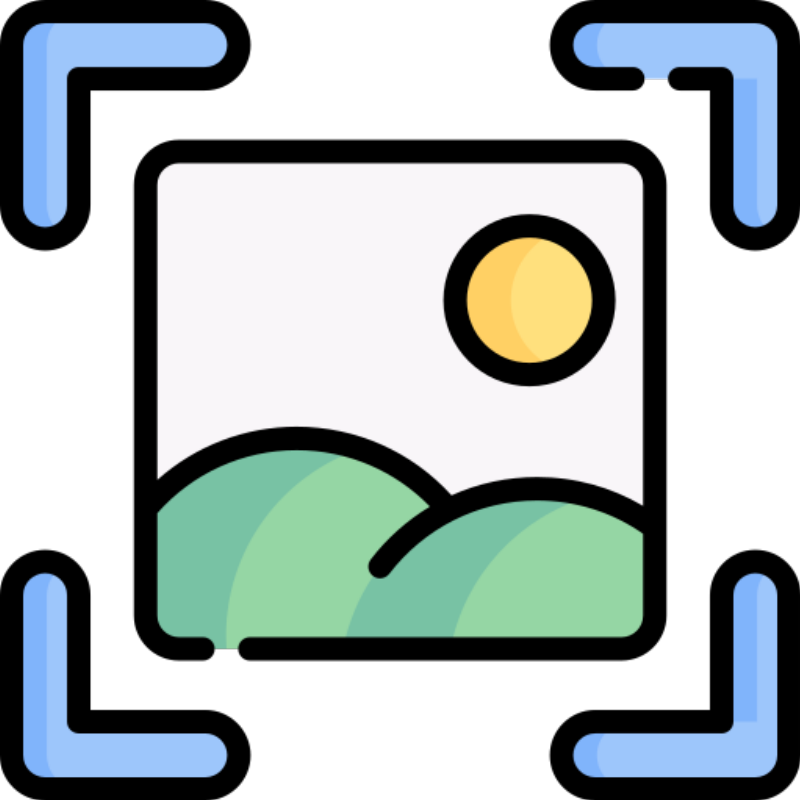}
& \hlred{Image} \newline \includegraphics[width=0.1\textwidth]{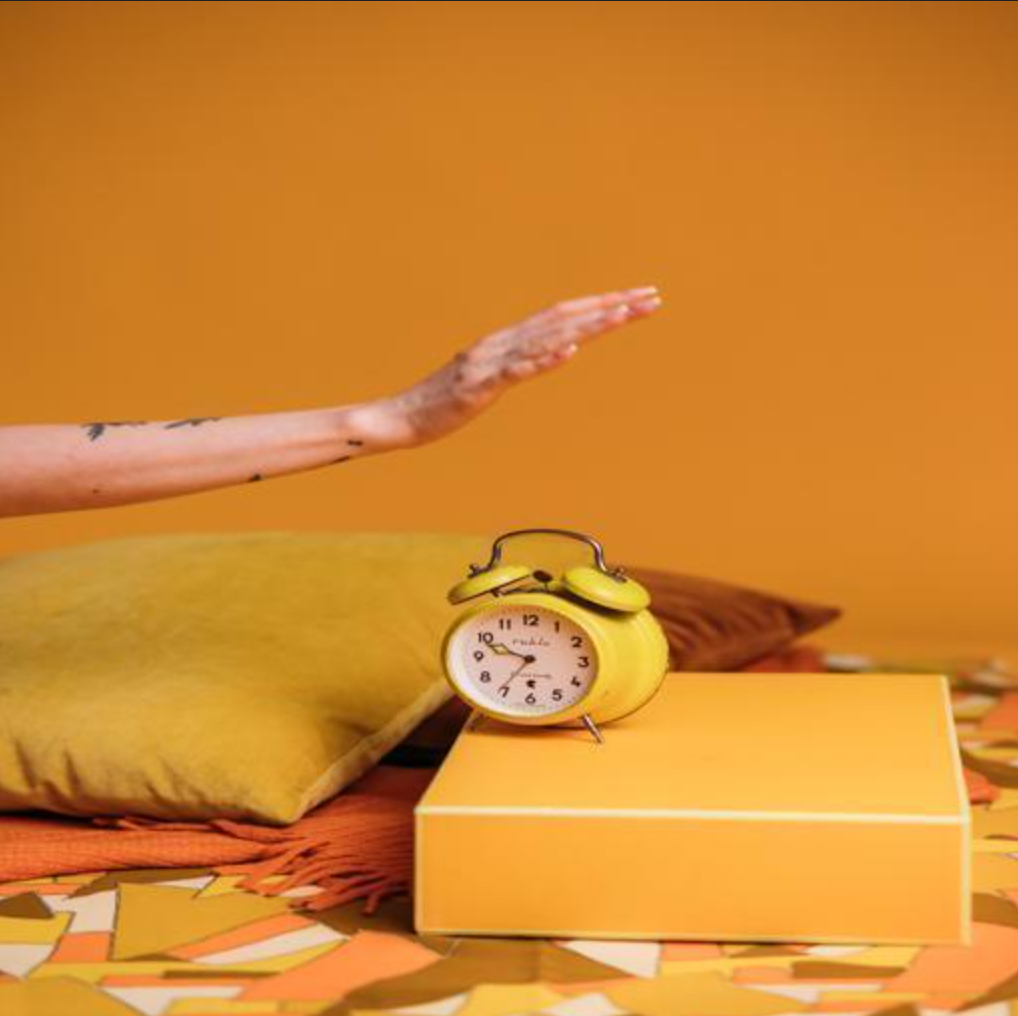}
& \hlgreen{Target Entity} \newline \texttt{a yellow alarm clock}
& \hlyellow{\texttt{[219,261,337,370]}} \newline   \includegraphics[width=0.1\textwidth]{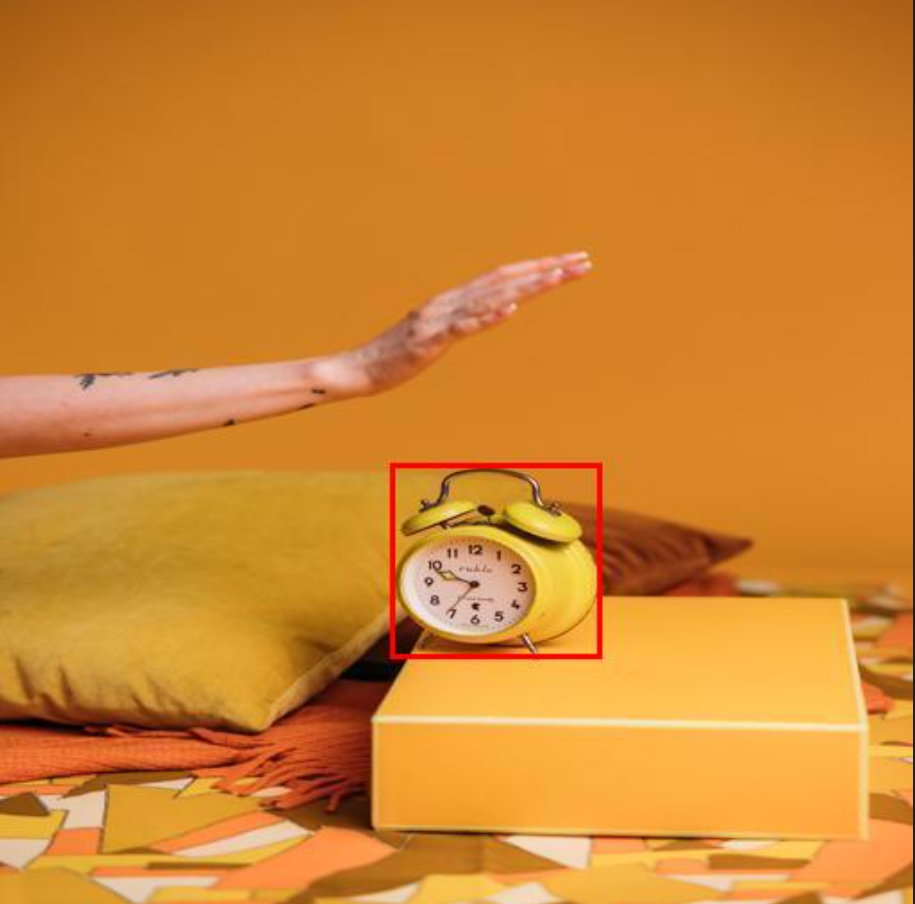}%\texttt{[[105, 240, 144, 480]]}
%with Framed Target Entity
&& Obtain the coordinates of regions from the \hlred{Image} corresponding to the \hlgreen{Target Entity}.\\
\midrule
\textbf{\textsf{Highlight}} \newline \includegraphics[width=0.08\textwidth]{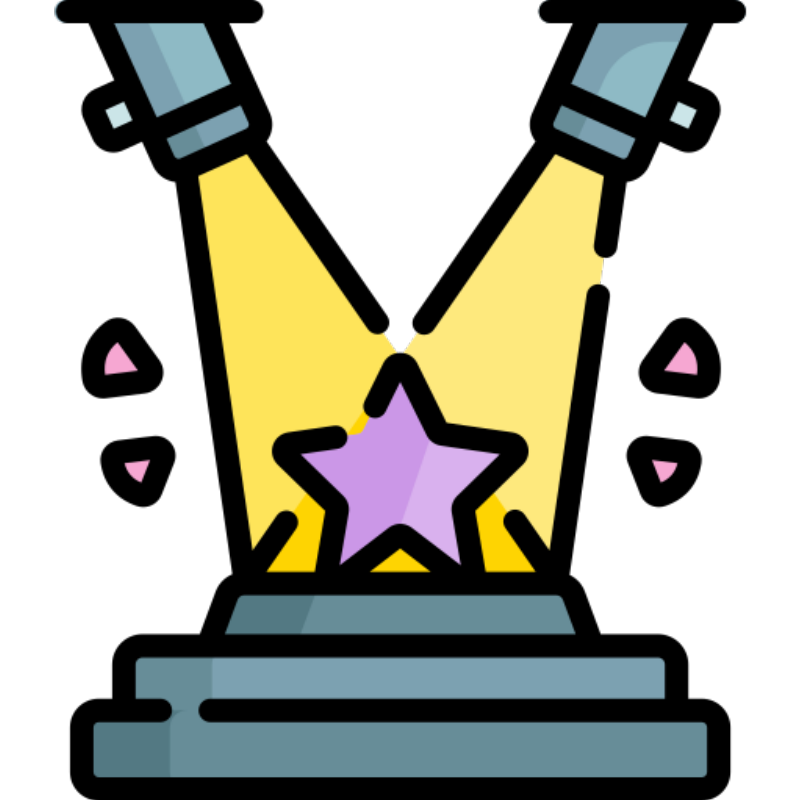}
& \hlred{Image} \newline \includegraphics[width=0.1\textwidth]{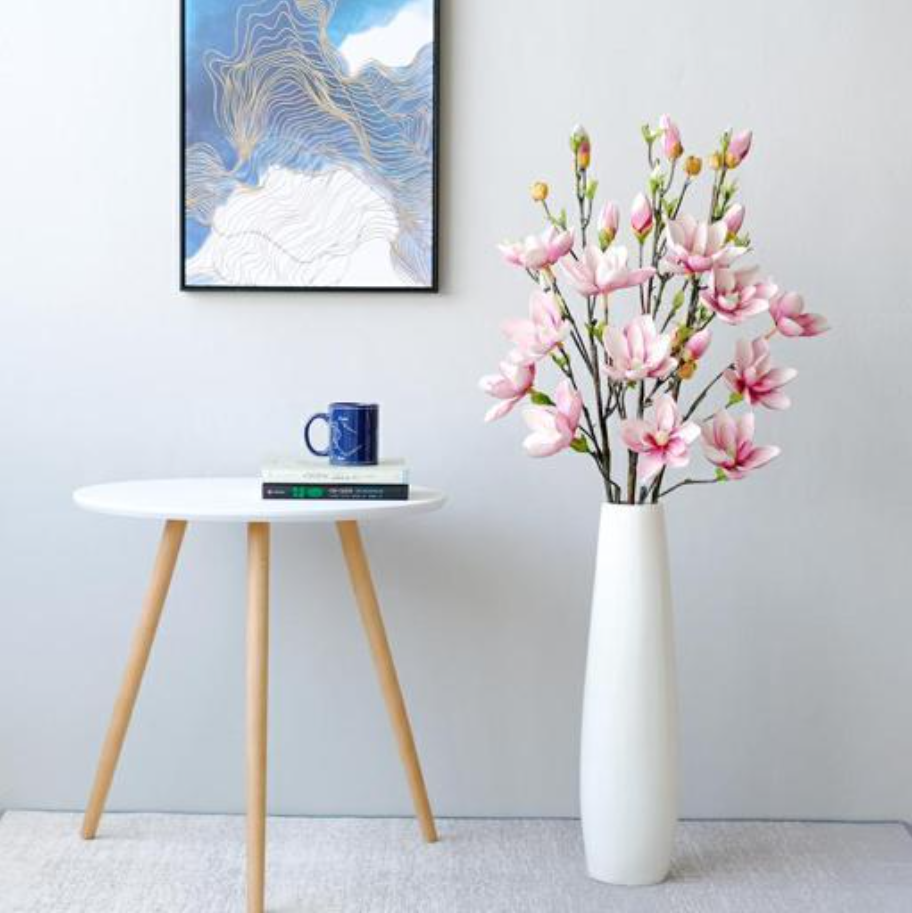}
& \hlblack{Region} \newline \texttt{[324,281,381,497]} & \hlyellow{Edited Image} \newline \includegraphics[width=0.1\textwidth]{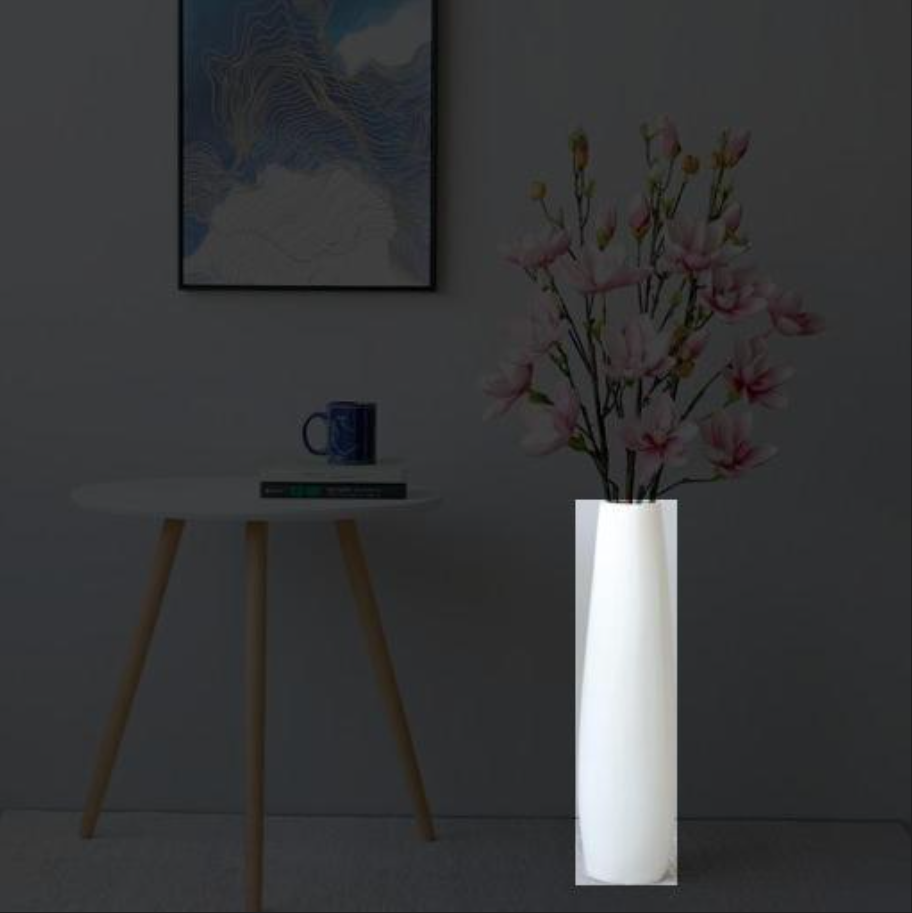}
&& Highlight the listed \hlblack{Region} in the \hlred{Image}.
\\
\midrule
\textbf{\textsf{Difference}} \newline \includegraphics[width=0.1\textwidth]{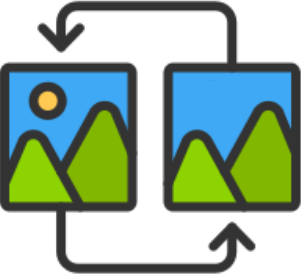}
& \hlred{Image 1} \newline \includegraphics[width=0.1\textwidth]{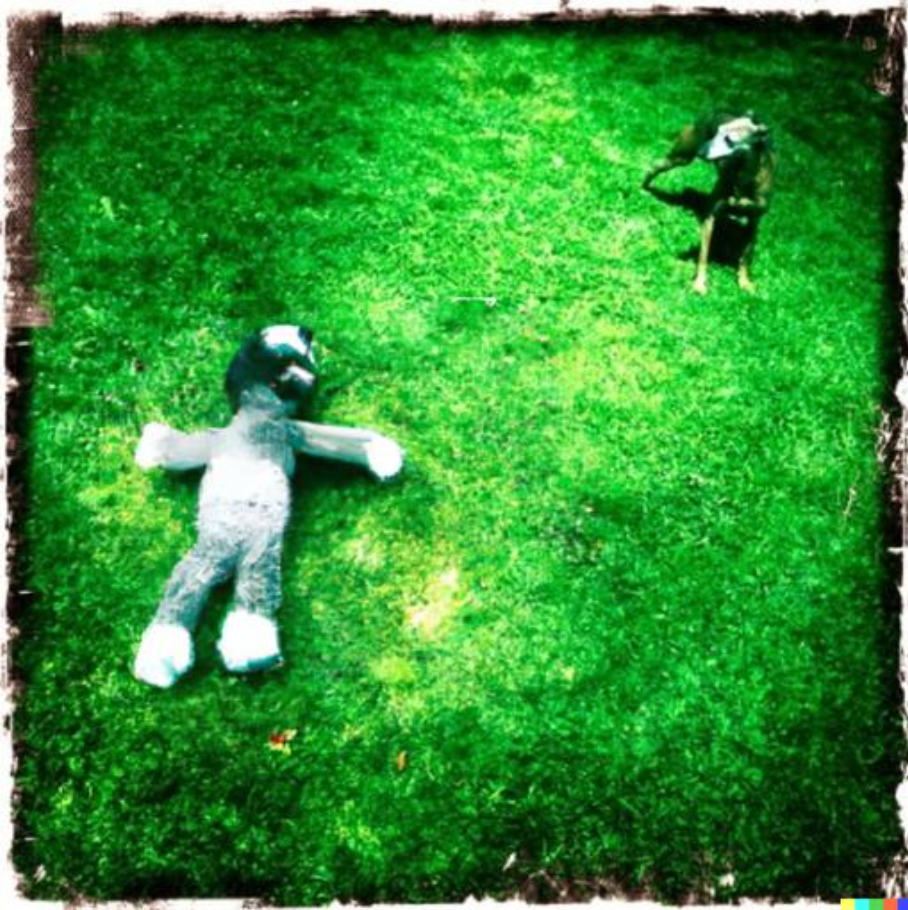}
& \hlgreen{Image 2} \newline \includegraphics[width=0.1\textwidth]{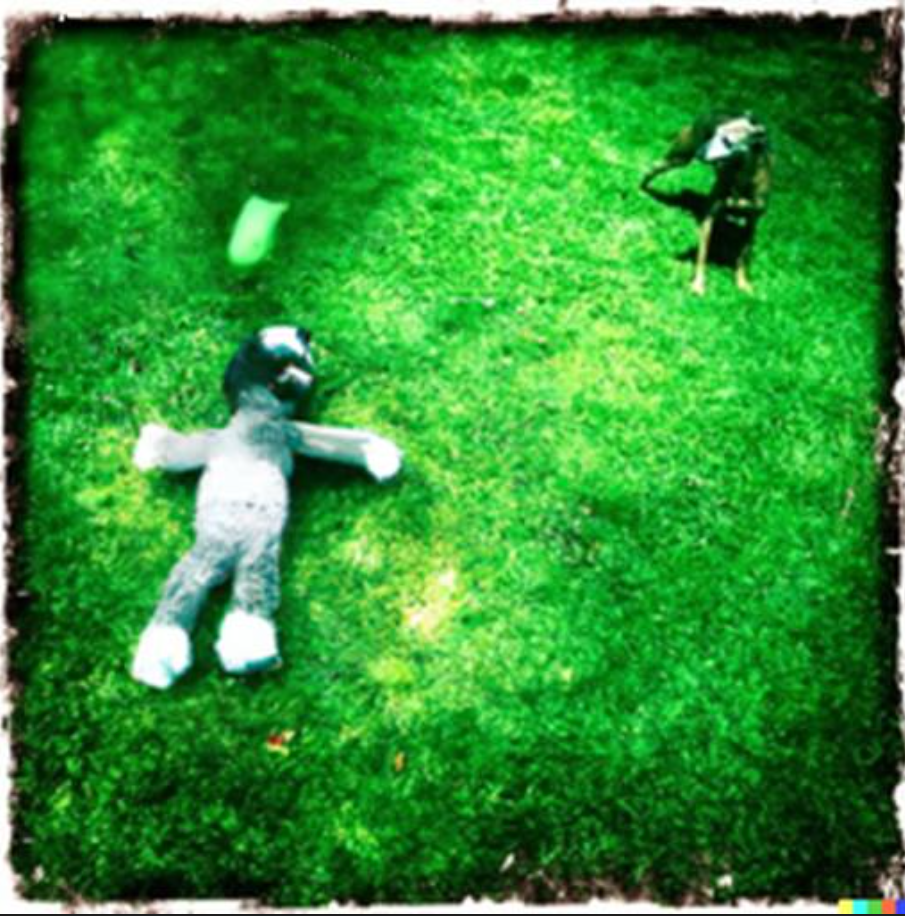} 
& \hlyellow{\texttt{[128,109,164,150]}} \newline \includegraphics[width=0.22\textwidth]{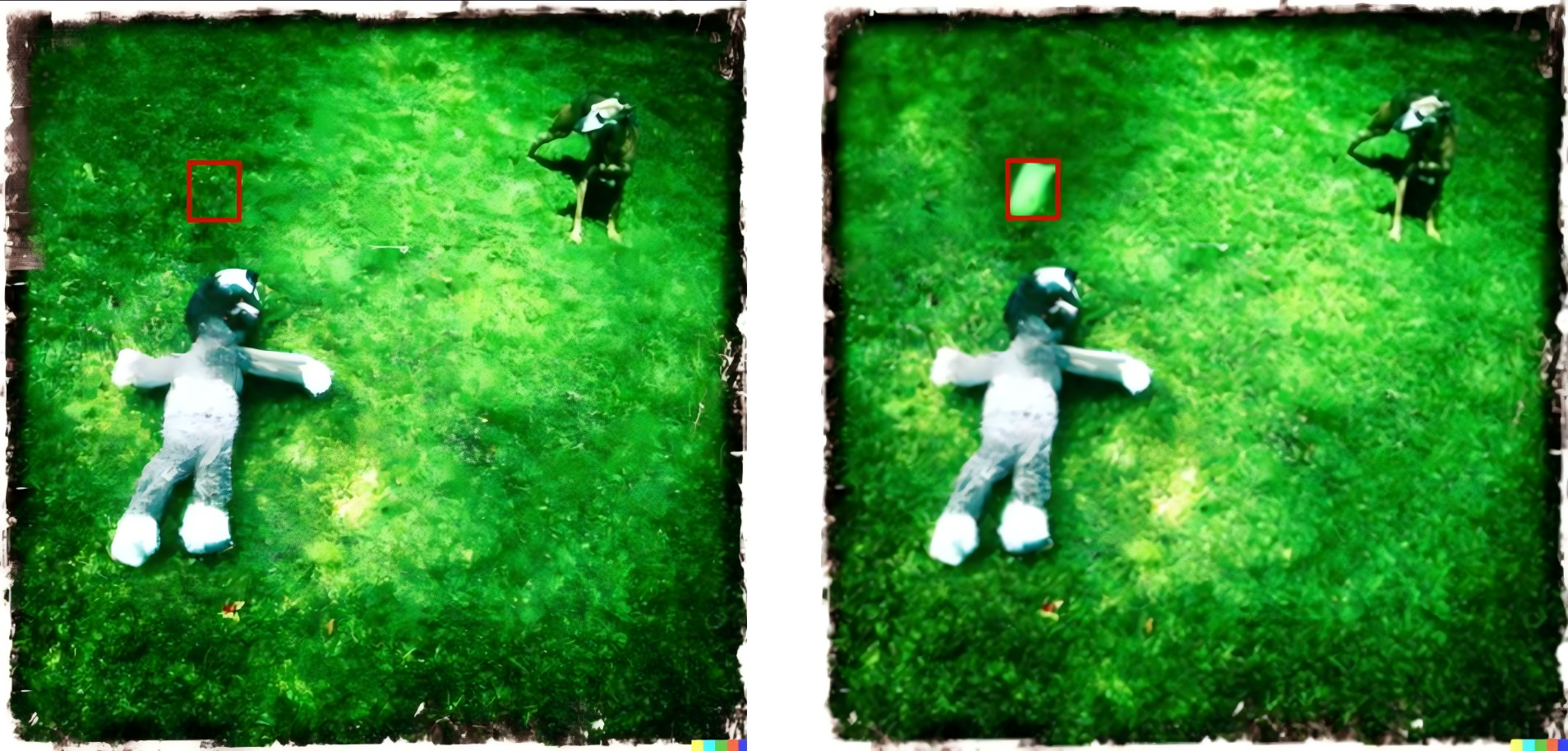}
&
& Identify the pixel difference between \hlred{Image 1} and \hlgreen{Image 2}.
\\
\midrule
\textbf{\textsf{Scene Graph}} \newline \includegraphics[width=0.08\textwidth]{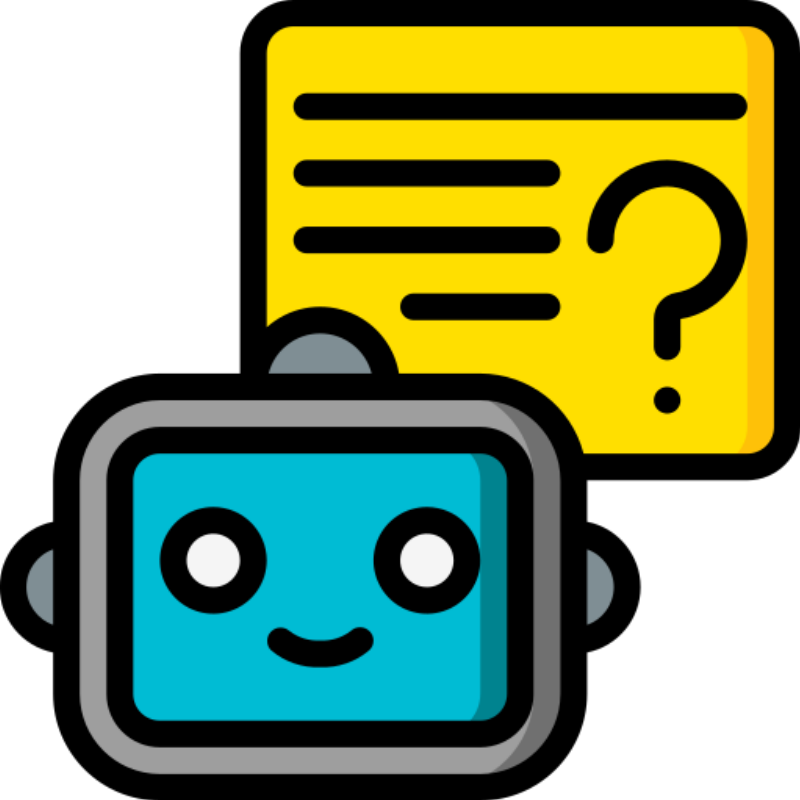}
&\hlred{Image} \newline \includegraphics[width=0.1\textwidth]{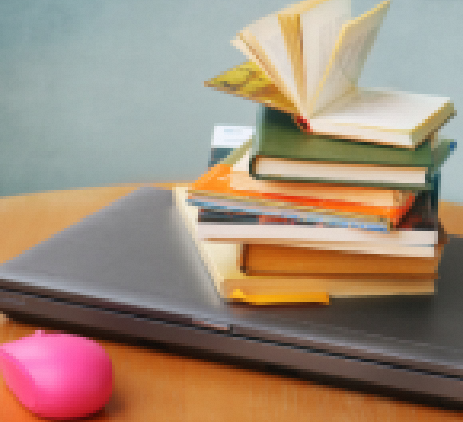}
&
&\hlyellow{A dict about objects and attributes}  \newline \includegraphics[width=0.17\textwidth]{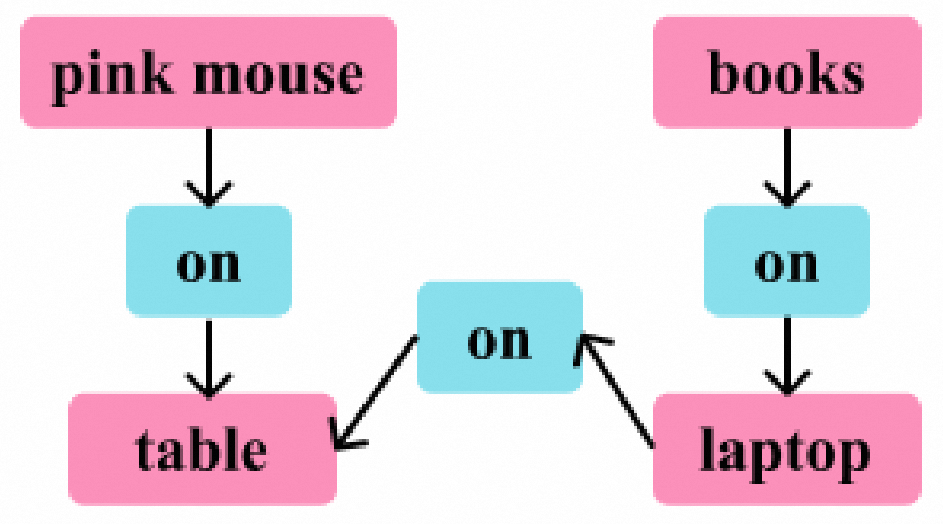}
&& 
Analyzed by LMMs, a structured description of the objects, their attributes, and the relationships in \hlred{Image}.
\\
\bottomrule
\end{tabular}
\end{adjustbox}
\caption{Tools used in our \score framework.}
\label{tab:tool}
\end{table*}

Evaluating image generation that involves multiple conditions can be challenging.
Drawing inspiration from prior research~\cite{cheng-etal-2024-least,sgedit}
we have developed a multi-functional toolbox, including \texttt{Grounding}, \texttt{Difference}, \texttt{Highlight} and \texttt{Scene Graph}.
Each tool is designed to target specific aspects of image analysis or editing,
and outputs either a modified image or textual information. 
Detailed descriptions of each tool can be found in Table~\ref{tab:tool}.

Specifically, we implement \texttt{Grounding} with GroundingDino~\cite{grounding_dino}.
\texttt{Scene Graph} uses the same prompting method as CCoT~\cite{ccot} based on GPT-4o. 
This tool can also function effectively with other open-source LMMs (refer to Sec.~\ref{sec:corr_study}). 
To assist LMMs in detecting subtle differences between edited images, the \texttt{Difference} tool compares the pixels of two images and identifies the locations of the variations.
The \texttt{Highlight} tool emphasizes selected regions by reducing the pixel values of areas outside the highlighted zone to $1/4$ of their original values, thereby darkening these areas and accentuating the highlighted region.
This tool is typically used after the \texttt{Grounding} and \texttt{Difference} tools have provided the region coordinates.

\begin{figure}[!t]
    \centering
    \includegraphics[width=1\linewidth]{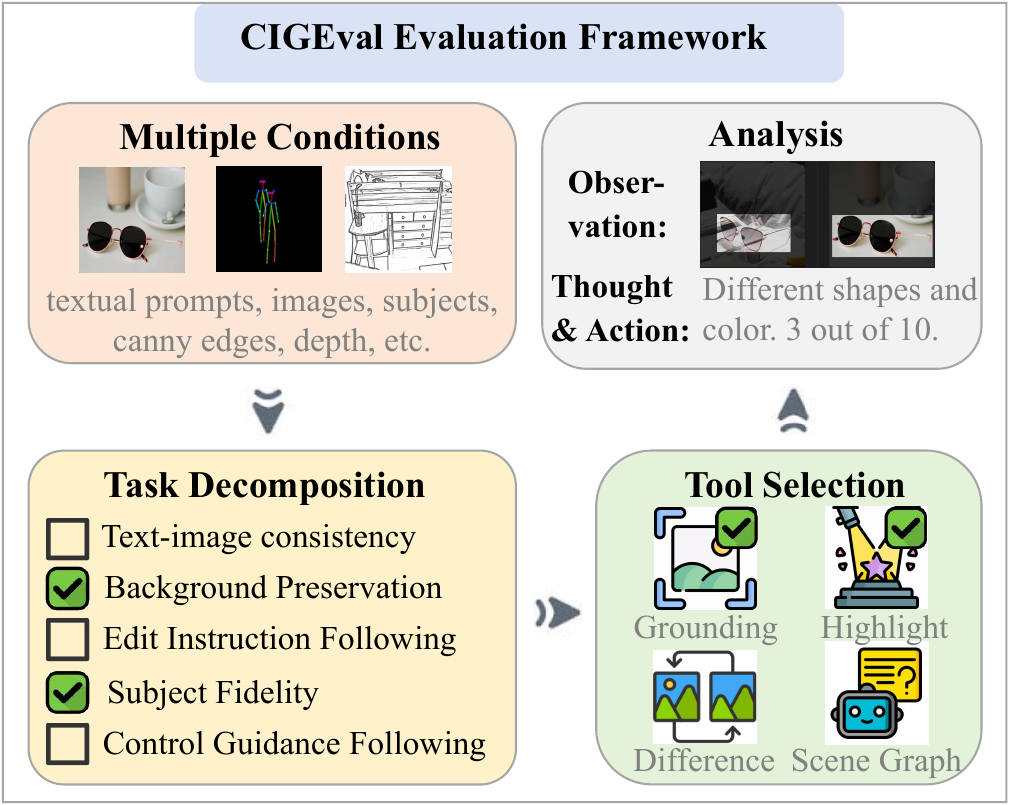}
    \caption{The evaluation process of \score regarding the example in Figure~\ref{fig:intro}. \score autonomously selects appropriate tools for each decomposed sub-task, and then conducts fine-grained analyses based on the observed tool outputs.}
    \label{fig:method_main}
\end{figure}

\subsection{Framework}
\label{sec:framework}

In our approach, we conceptualize the image evaluation process as an agent task. 
As shown in Figure~\ref{fig:method_main}, 
the core of \score is a well-instructed LMM,
which autonomously utilizes tools to assess a wide range of conditional image generation tasks. 
The prompts used in this framework are listed in Appendix~\ref{sec:appendix_prompt}.

Specifically,
we adopt a divide-and-conquer scheme for evaluating images generated under multiple conditions.
For example, in the subject-driven image editing task in Figure~\ref{fig:intro},
the desired synthesized image will incorporate the object from a subject reference while maintaining the background of the source image.
Therefore, we break down each evaluation task into several fine-grained sub-questions from the listed below:
(1) Is the image generation following the prompt?
(2) Is the image editing following the instruction?
(3) Is the image performing minimal edit without changing the background?
(4) Is the object in the Image following the provided subject?
(5) Is the image following the control guidance?
Then, for each sub-question,
\score selects the most suitable tool from its toolbox, focusing on the specific aspect of evaluation. 
For example, \texttt{Grounding} and \texttt{Highlight} are utilized for comparing specific regions of the image, while \texttt{Scene Graph} evaluates background preservation and the extent of over-editing.
With these intermediate results, the LMM analyzes the tool outputs and assigns scores in the ReAct format \cite{yao2023react}, ranging from 0 to 10, which are normalized to the [0.0, 1.0] range for comparison with human ratings.
These fine-grained scores are aggregated through:
\begin{equation}
    O = \min(\alpha_1, ..., \alpha_i)
\end{equation}
where $\alpha_i$ represents one of the sub-scores.
In accordance with the setting of \citet{ku2023imagenhub},
we assume each sub-score weights the same and used $\min$ operation to emphasize the importance of meeting all criteria without exception.

\subsection{Agent Tuning}
\label{sec:tuning}

Previous research has primarily relied on closed-source LMMs to address agentic tasks,
mainly due to their superior abilities in tool calling and instruction following~\cite{chen-etal-2024-agent,song-etal-2024-agentbank,zeng-etal-2024-agenttuning,xu2023reasoning}.
As evidenced in Table~\ref{tab:result_main},
open-source models significantly underperform compared to GPT-4o. 
To bridge this gap and
empower smaller LMMs as effective evaluators, 
we aim to perform supervised fine-tuning on 7B models
to integrate agentic capabilities into them.

To curate high-quality trajectory data,
we employ GPT-4o to carry out the evaluation process in Section~\ref{sec:framework}.
The process begins by providing GPT-4o with evaluation instructions and corresponding images. 
At each turn, 
the agent receives an \emph{observation}, formulates plans and thoughts as \textit{thought}, and invokes relevant tools through \textit{action}. 
The tool outputs serve as new observations for the subsequent turn. 
By iterating the above process, we can construct a complete evaluation trajectory including the initial instruction, intermediate steps (i.e., observations, thoughts, actions), and the final scoring result. 
To ensure the quality of these trajectories, 
we exclude samples where the discrepancy between predicted scores and human evaluation scores exceeds 0.3.
Using 60\% of the ImagenHub data, we ultimately gather 2,274 high-quality trajectories for supervised fine-tuning.

Using this structured trajectory data, we perform supervised fine-tuning on Qwen2-VL-7B-Instruct and Qwen2.5-VL-7B-Instruct \cite{wang2024qwen2vl}.
Formally, each sample's evaluation trajectory is represented as $\langle o_0, t_1, a_1, \ldots, o_{n-1}, t_n, a_n, o_n \rangle$, 
where $o_i$, $t_i$, and $a_i$ denote the observation, {thought}, and {action} at each turn ($i$) respectively. 
Specifically, $o_0$ refers to the initial observation consisting of the evaluation instructions and accompanied images, and $o_n$ denotes the final score.
At each turn, based on the preceding trajectory $c_i = \langle o_0, t_1, a_1, \ldots, o_{i-1} \rangle$, 
the agent aims to generate thought $t_i$ and action $a_i$.
During the fine-tuning process, we only compute the cross-entropy loss for $t_i$ and $a_i$ while $c_i$ is masked:
\begin{equation}
    \mathcal{L} = -\log \sum_{i=1}^{n} \text{Pr}(t_i,a_i|c_i).
\end{equation}

\section{Experiments}

\subsection{Evaluation Benchmark}
ImagenHub \citep{ku2023imagenhub} is a standardized benchmark for evaluating conditional image generation models with human raters. 
The statistics for ImagenHub are presented in Table~\ref{tab:imagenhub_data_info}.
This large-scale benchmark covers 7 mainstream tasks, 29 models, 4.8K synthesized images, and 14.4K human ratings,
making it suitable for calculating correlation scores between automatic evaluation metrics and human raters.
A list of 29 evaluated models can be found in Appendix~\ref{sec:appen_imagenhub}.

Each image was assessed by three human raters according to the guidelines of the defined task, and a final score in the range [0.0, 1.0] was reported for the average score.
Images are scored in two aspects: 
(1) Semantic Consistency assesses how well the generated image aligns with the given conditions, such as prompts and subject tokens, ensuring coherence and relevance to the specified task criteria.
(2) Perceptual Quality evaluates the extent to which the generated image appears visually authentic and conveys a sense of naturalness. 
In this work, we focus on the Semantic Consistency score, leaving the exploration of Perceptual Quality for future research.

\begin{table}[t]
\centering
\scalebox{0.85}{
\begin{tabular}{c|c|c}
\toprule
\textbf{\# Instructions} & \textbf{\# Images} & \textbf{\# Human Ratings} \\
\midrule
\multicolumn{3}{c}{\textbf{Text-guided Image Generation (5 models)}} \\
\midrule
197 & 985 & 2955 \\
\midrule
\multicolumn{3}{c}{\textbf{Mask-guided Image Editing (4 models)}} \\
\midrule
179 & 716 & 2148 \\
\midrule
\multicolumn{3}{c}{\textbf{Text-guided Image Editing (8 models)}} \\
\midrule
179 & 1432 & 4296 \\
\midrule
\multicolumn{3}{c}{\textbf{Subject-driven Image Generation (4 models)}} \\
\midrule
150 & 600 & 1800 \\
\midrule
\multicolumn{3}{c}{\textbf{Subject-driven Image Editing (3 models)}} \\
\midrule
154 & 462 & 1386 \\
\midrule
\multicolumn{3}{c}{\textbf{Multi-concept Image Composition (3 models)}} \\
\midrule
102 & 306 & 918 \\
\midrule
\multicolumn{3}{c}{\textbf{Control-guided Image Generation (2 models)}} \\
\midrule
150 & 300 & 900 \\
\midrule
\multicolumn{3}{c}{\textbf{Sum of 7 tasks}} \\
% & & Sum of 7 tasks \\
\midrule
1111 & 4801 & 14403 \\
\bottomrule
\end{tabular} }
\caption{Statistics of ImagenHub: the number of instructions, evaluated models, synthesized images, and human ratings used in this paper.}
\label{tab:imagenhub_data_info}
\end{table}

\subsection{Existing Auto-metrics}
\label{sec:auto_metric_survey}
Here we list some prominent automatic metrics:
\begin{itemize}
    \item \textbf{CLIP-Score} \citep{hessel2021clipscore}: This metric computes the average cosine similarities between prompt and generated image CLIP embeddings, making it a popular choice for assessing image-text alignment.
    \item \textbf{LPIPS} \citep{zhang2018perceptual} measures the similarity between two images in a manner that aligns with human perception. 
    \item \textbf{CLIP-I} \citep{gal2022image} calculates the average pairwise cosine similarities between CLIP embeddings of generated and source images.
    \item \textbf{DINO} \citep{ruiz2023dreambooth} is computed by the mean cosine similarity computed between the DINO embeddings of ViT-S/16 \citep{Caron2021EmergingPI} for both synthesized and source images.
    \item \textbf{\textsc{VIEScore}} \citep{ku-etal-2024-viescore} prompts large multimodal models to evaluate generated images in an explainable and fine-grained manner. Based on GPT-4o, it currently represents the state-of-the-art across all seven tasks on ImagenHub.
\end{itemize}

\begin{table*}[!ht]
\centering
\scalebox{0.85}{
\begin{tabular}{
    >{\centering\arraybackslash}m{2.2cm}|
    >{\centering\arraybackslash}m{1.8cm}
    >{\centering\arraybackslash}m{1.6cm}
    >{\centering\arraybackslash}m{1.6cm}
    >{\centering\arraybackslash}m{1.8cm}
    >{\centering\arraybackslash}m{1.6cm}
    >{\centering\arraybackslash}m{1.6cm}
    >{\centering\arraybackslash}m{1.8cm}|
    >{\centering\arraybackslash}m{0.9cm}}
\toprule
\textbf{Method} & \textbf{Text-guided IG} & \textbf{Mask-guided IE} & \textbf{Text-guided IE} & \textbf{Control-guided IG} & \textbf{Subject-driven IG} & \textbf{Subject-driven IE} & \textbf{Multi-concept IC} & \textbf{Avg.} \\
\midrule
Human Raters & 0.5044 & 0.5390 & 0.4230 & 0.5443 & 0.4780 & 0.4887 & 0.5927 & 0.4700 \\
\midrule
CLIPScore & -0.0817 & - & - & - & - & - & - & - \\
LPIPS & - & -0.1012 & 0.0956 & 0.3699 & - & - & - & - \\
DINO & - & - & - & - & 0.4160 & 0.3022 & 0.0979 & - \\
CLIP-I & - & - & - & - & 0.2961 & 0.2834 & 0.1512 & - \\
\midrule
\multicolumn{9}{c}{\textit{LLaMA3-LLaVA-NeXT-8B}} \\
\textsc{VIEScore} & 0.1948 & 0.2037 & 0.0363 & 0.4001 & 0.1592 & -0.1153 & 0.1308 & 0.1432 \\
\rowcolor[HTML]{EFEFEF} 
\score & 0.1420 & 0.2843 & 0.0744 & 0.4487 & 0.2891 & -0.0699 & 0.3704 & 0.2164 \\
\midrule
\multicolumn{9}{c}{\textit{Qwen2.5-VL-7B-Instruct}} \\
\textsc{VIEScore} & 0.4218 & 0.3555 & 0.0252 & 0.2836 & 0.4264 & -0.0452 & 0.3328 & 0.2516 \\
\rowcolor[HTML]{EFEFEF} 
\score & 0.4347 & 0.4685 & 0.2567 & 0.3752 & 0.4374 & 0.4863 & 0.3251 & 0.3780 \\
\midrule
\multicolumn{9}{c}{\textit{GPT-4o}} \\
\textsc{VIEScore} & 0.4989 & 0.5421 & 0.4062 & 0.4972 & 0.4806 & 0.4800 & 0.4516 & 0.4459 \\
\rowcolor[HTML]{EFEFEF} 
\score & \textbf{0.5027} & \textbf{0.5465} & \textbf{0.4090} & \textbf{0.5402} & \textbf{0.4930} & \textbf{0.5185} & \textbf{0.4931} & \textbf{0.4625} \\
\bottomrule
\end{tabular}}
\caption{Spearman correlation scores across 7 conditional image generation tasks with different LMMs as backbone. The abbreviations ``IG'', ``IE'' and ``IC'' stand for ``Image Generation'', ``Image Editing'' and ``Image Composition'' respectively. ``-'' means not applicable.}
\label{tab:result_main}
\end{table*}

\subsection{Implementation Details}

In all experiments, GPT-4o refers to the model version \texttt{GPT-4o-2024-05-13}, aligning with the original \textsc{VIEScore} paper~\cite{ku-etal-2024-viescore}. 
For the experiments in Sec.~\ref{sec:corr_study},
we evaluate using the entire ImagenHub benchmark. 
In the ablation study, we randomly select 50 images for each task. 
For the experiments in Sec.~\ref{sec:tuning_exp},
we generate training data using 60\% of the ImagenHub dataset, as described in Section~\ref{sec:tuning}, and use the remaining data for testing. 
We fine-tune the \texttt{Qwen2-VL-7B-Instruct} and \texttt{Qwen2.5-VL-7B-Instruct} models using Megatron-LM.
The fine-tuning process employs a learning rate of 1e-5 and a batch size of 128, with a sequence length of 32,768. 
We use AdamW optimizer with a cosine learning scheduler with 3\% warm-up steps.

\subsection{Main Experiments}
\label{sec:corr_study}

For all presented correlations, we applied Fisher
Z-transformation to estimate the average Spearman
correlation$\in[-1,1]$ across models and tasks.

\noindent \textbf{Metric-to-Human (M-H) correlations.} 
In Table \ref{tab:result_main}, we present the correlations across all tasks, utilizing different backbone models. 
When using GPT-4o as the underlying LMM, \score achieves the state-of-the-art performance across all 7 tasks. It achieves an average Spearman correlation of 0.4625 with human raters, closely matching the human-to-human correlation. The primary improvements are observed in tasks involving multiple conditions, such as control-guided image generation and multi-concept image composition, where previous evaluation metrics struggle.

When the underlying LMM is replaced with different open-source models, \score consistently outperforms \textsc{VIEScore}.  
However, the performance of open-source models is still poor and falls significantly behind GPT-4o. Therefore, we perform agent tuning on these models as described in Sec.~\ref{sec:tuning} and report their improved performance in Sec.~\ref{sec:tuning_exp}.
Overall, the experiment demonstrates that \score outperforms \textsc{VIEScore} across a variety of image editing and generation tasks, consistently maintaining its edge even when different underlying LMMs are used.

\begin{table}[t]
    \centering
    \scalebox{0.85}{
    \begin{tabular}{lc}
        \toprule
        \textbf{Configuration} & \textbf{Avg.} \\
        \midrule
        \rowcolor[HTML]{EFEFEF} 
        \textbf{\score} & 0.7262 \\
        w/o \texttt{Grounding} & 0.6376 \\
        w/o \texttt{Difference} & 0.7020 \\
        w/o \texttt{Scene Graph} & 0.6471 \\
        \midrule
        \texttt{Scene Graph} with \textit{Qwen2.5-VL-7B-Instruct} & 0.7120 \\
        \texttt{Scene Graph} with \textit{Qwen2.5-VL-70B-Instruct} & 0.7311 \\
        \bottomrule
    \end{tabular}}
    \caption{Ablation study regarding tools in \score (GPT-4o) with different configurations.}
    \label{tab:ablation}
\end{table}

\noindent \textbf{Ablation study.}
To assess the necessity of each tool in \score, we conducted an ablation study detailed in Table \ref{tab:ablation}. 
Since \texttt{Highlight} is often accompanied with \texttt{Grounding} and \texttt{Difference},
we do not perform specific ablation on \texttt{Highlight}.
The study shows that the complete \score configuration achieves the highest average score of 0.7262. 
When each tool is removed, a noticeable drop is observed, underscoring the necessity of every tool for effective performance. 
 
On the other hand, when the implementation of the \texttt{Scene Graph} was switched from GPT-4o to an open-source model, the evaluation results remained high with minimal variation. In fact, when replaced with the Qwen2.5-VL-70B, the performance improved further, showcasing the robustness of our agentic framework.
Overall, the ablation study underscores that each tool in the \score configuration is useful, and their collective integration is crucial for achieving superior performance.

\begin{table*}[!ht]
\centering
\scalebox{0.85}{
\begin{tabular}{
    p{2.4cm}|
    >{\centering\arraybackslash}m{1.7cm}
    >{\centering\arraybackslash}m{1.6cm}
    >{\centering\arraybackslash}m{1.6cm}
    >{\centering\arraybackslash}m{1.8cm}
    >{\centering\arraybackslash}m{1.6cm}
    >{\centering\arraybackslash}m{1.6cm}
    >{\centering\arraybackslash}m{1.8cm}|
    >{\centering\arraybackslash}m{0.9cm}}
\toprule
\textbf{Method} & \textbf{Text-guided IG} & \textbf{Mask-guided IE} & \textbf{Text-guided IE} & \textbf{Control-guided IG} & \textbf{Subject-driven IG} & \textbf{Subject-driven IE} & \textbf{Multi-concept IC} & \textbf{Avg.} \\
\midrule
Previous SOTA & 0.3081 & 0.3167 & 0.4649 & 0.5246 & {0.7105} & 0.4694 & 0.5616 & 0.4458 \\
\midrule
\multicolumn{9}{c}{\textit{Qwen2.5-VL-7B-Instruct}} \\
\textsc{VIEScore} & 0.3457 & 0.0158 & 0.0086 & 0.2395 & 0.1837 & 0.0967 & 0.4388 & 0.1876 \\
\score & 0.1890 & 0.1418 & 0.4586 & 0.3130 & 0.4485 & 0.5216 & 0.4496 & 0.3455 \\
\rowcolor[HTML]{EFEFEF} 
\textbf{~~+ Tuning} & {0.4609} & 0.2796 & {0.5916} & 0.5876 & 0.4659 & 0.5458 & 0.5778 & 0.4631 \\
\midrule
\multicolumn{9}{c}{\textit{Qwen2-VL-7B-Instruct}} \\
\textsc{VIEScore} & 0.3699 & -0.1398 & 0.1024 & 0.3420 & 0.1553 & 0.0682 & 0.5129 & 0.1989 \\
\score & 0.3054 & 0.1974 & 0.1438 & 0.2615 & 0.5096 & 0.1226 & 0.5035 & 0.2840 \\
\rowcolor[HTML]{EFEFEF} 
\textbf{~~+ Tuning} & 0.4099 & {0.5272} & 0.3846 & {0.6096} & 0.6445 & {0.5975} & {0.6691} & \textbf{0.4997} \\
\bottomrule
\end{tabular}}
\caption{Spearman correlations across 7 tasks with \textsc{VIEScore} and \score based on open-source small LMMs. ``Previous SOTA'' here means \textsc{VIEScore} based on GPT-4o.}
\label{tab:agent_main}
\end{table*}

\subsection{\score with Agent Tuning}
\label{sec:tuning_exp}
The experimental results in Table \ref{tab:agent_main} show the performance of \score after agent tuning. 
Despite utilizing 7B open-source LMMs as the underlying model, Qwen2-VL-7B-Instruct and Qwen2.5-VL-7B-Instruct demonstrate a 76\% and 34\% improvement in correlation after fine-tuning, respectively,
With only 2,274 filtered evaluation trajectories, 
the fine-tuned 7B models surpass the previous state-of-the-art VIEScore based on GPT-4o.
This demonstrates the data efficiency of agent tuning and the importance of synthetic data quality.

\subsection{Case Study}

\begin{figure*}[tb]
	\centering
        \small
	\includegraphics[width=1.00\textwidth]{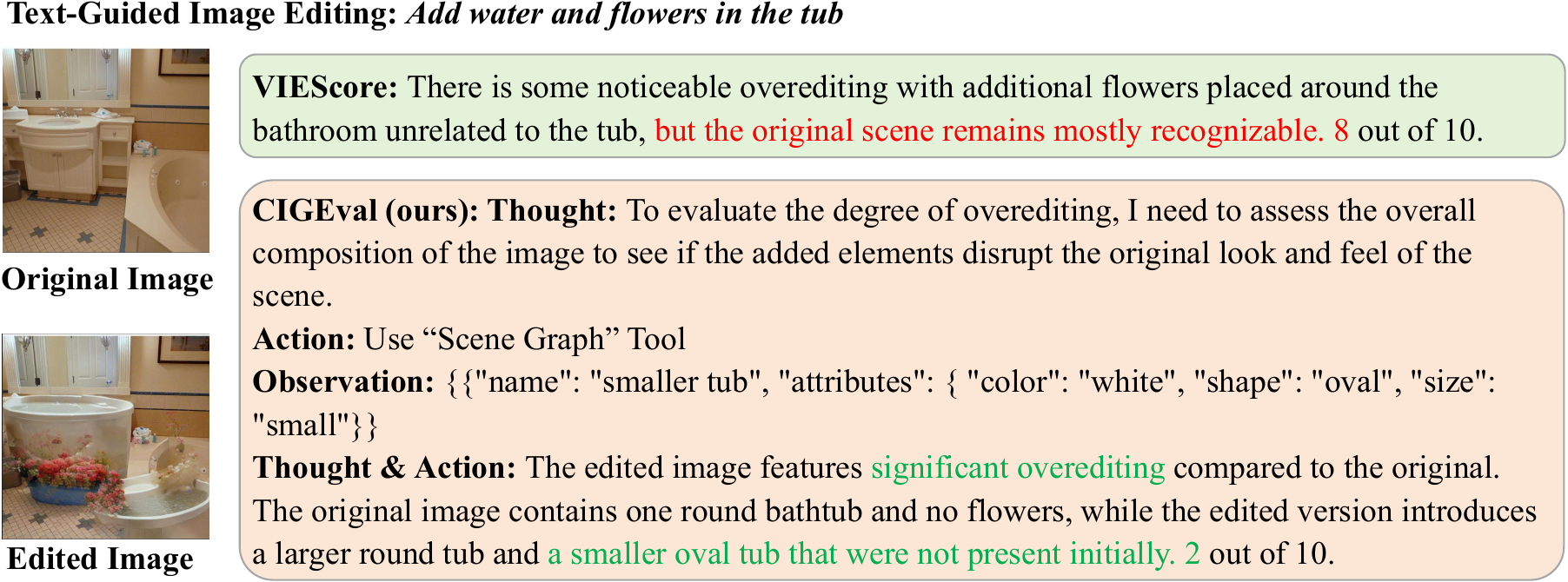}
	\caption{Case study of a text-guided image editing example with a low human annotation score.}
	\label{fig: case study}
\end{figure*}

To demonstrate the effectiveness of our \score framework and the importance of each tool,
we present a subject-driven image editing example in Figure~\ref{fig:intro}, a text-guided image editing example in Figure~\ref{fig: case study}, and a multi-concept image composition example in Figure~\ref{fig:multi_comp_eg}.
In the first and third example, by directly prompting in \textsc{VIEScore},
GPT-4o struggles to compare the similarity of specific objects between two images.
By grounding and highlighting the focused object (i.e., glasses and flowers),
GPT-4o can find the difference in shapes and colors within our framework.
In the second example,
when discussing the background preservation aspect,
\textsc{VIEScore} considers the over-editing small.
However, in our framework, \score first calls the \texttt{Scene Graph} tool
to get an overall composition of the edited image,
and then finds out a newly-added tub based on tool outputs,
successfully arriving at the correct score.
These examples have shown \score's ability to autonomously select appropriate tools and reach correct conclusions based on the observation,
which makes \score a better evaluator than VIEScore.

\begin{figure*}[t]
	\centering
	\includegraphics[width=1.00\textwidth]{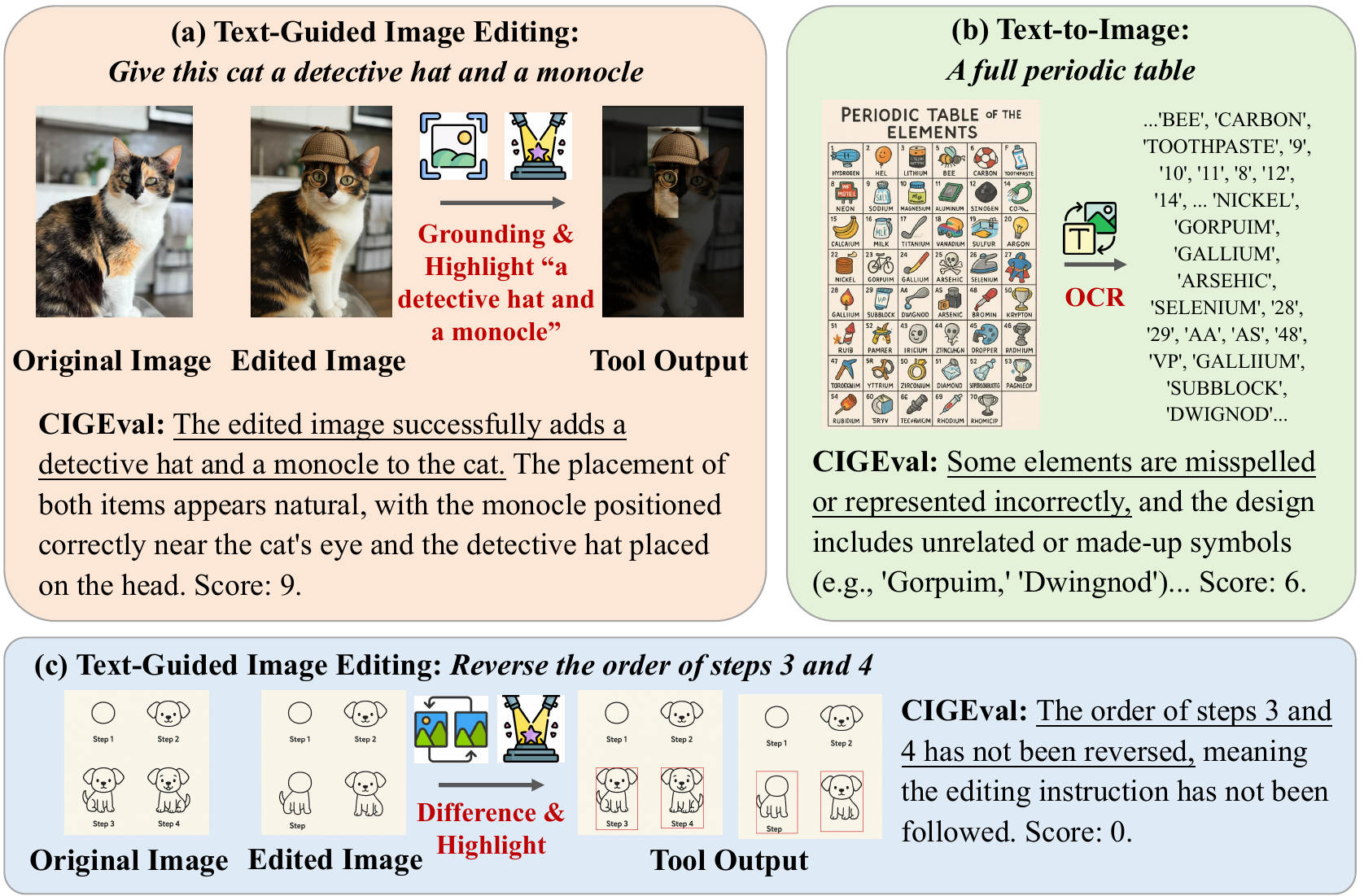}
	\caption{Case study of GPT-4o's image generation. Examples are adapted from OpenAI's official website.}
	\label{fig:4o_official}
\end{figure*}

\noindent \textbf{Preliminary study with GPT-4o image generation.}
Recently GPT-4o image generation has attracted wide attention.
As shown in Figure~\ref{fig:4o_official},
we extend \score with an additional OCR tool and find that our framework assigns appropriate scores to 4o-generated images on OpenAI's official website\footnote{\url{https://openai.com/index/introducing-4o-image-generation/}}.
Furthermore, we test GPT-4o image generation on ImagenHub's various tasks and report the CIGEval scores and human annotations (averaged between two annotators).
We have the following three findings:
(1) \score assigns scores that closely align with human annotations,
and effectively detects subtle flaws in GPT-4o-generated images.
(2) GPT-4o excels at tasks involving a single image as input, such as text-guided image generation and editing, as well as subject-driven image generation, as shown in Figures~\ref{fig:4o_sub_ig} to~\ref{fig:4o_text_ie}.
(3) GPT-4o struggles with complex tasks that require multiple images and control signals. 
For example, the subjects in Figures~\ref{fig:4o_multi_comp} and~\ref{fig:4o_sub_image_edit} are not accurately replicated, showing unintended changes in color or shapes.
Moreover, consistent with the findings of \citet{yan2025gptimgeval}, we observe that GPT-4o tends to favor a color palette dominated by yellow, orange, and warm lighting, as exemplified by the pot in Figure~\ref{fig:4o_multi_comp} and the man in the rearview mirror in Figure~\ref{fig:4o_text_ie}. 
Additionally, the control guidances (e.g., canny edges, OpenPose) are not strictly followed, as seen in Figure~\ref{fig:4o_condition}.

\section{Conclusion}

In this paper, we propose \score,
a unified, explainable and agentic framework
for image evaluation across seven popular conditional image evaluation tasks. 
\score utilizes large multimodal models (LMMs) at its core
to autonomously select tools for fine-grained evaluation.
Experiments show that, when using GPT-4o as the backbone model,
\score surpasses achieves a high correlation of 0.4625 with human raters,
closely matching the human-to-human correlation of 0.47.
Additionally, we have synthesized 2,274 high-quality evaluation trajectories to incorporate agentic capabilities into smaller LMMs. 
After agent tuning, the 7B LMMs surpass the performance of the previous state-of-the-art method based on the closed-source GPT-4o. 
These experimental findings and case studies on GPT-4o image generation 
suggest that \score holds substantial promise for replicating human-like performance in evaluating synthetic images.

\section*{Limitations}

Although \score improves the correlation between automatic image evaluators and human raters,
there are certain limitations to our approach. 
First, when using closed-source models' APIs, such as GPT-4o, there is a risk that AI-generated images resembling real people or photographs may be rejected by GPT-4o for evaluation, potentially affecting the framework's robustness. 
Second, our experiments primarily focus on evaluating images' consistency with multiple conditions, 
leaving the assessment of perceptual quality for future research. 
Due to the lack of a more comprehensive benchmark for conditional image generation,
we synthesized tuning data and conducted experiments exclusively on ImagenHub. 
Expanding our experiments to more text-to-image generation and text-based image editing datasets \citep{peng2024dreambench,hui2024hqedit} could be beneficial.
Finally, the training process currently utilizes only correct trajectory data and discards failed trajectory data. 
In the future, we aim to refine the \score framework to include a broader range of tasks and leverage failed data for contrastive training of the model.

% \section*{Potential Risks}
% Multimodal models can inadvertently perpetuate or amplify biases present in their training data. The interpretation and evaluation of synthetic images depend heavily on context. A multimodal model might not fully grasp certain images' nuances or cultural sensitivities, leading to inappropriate or offensive outputs.

\bibliography{0_main}

\appendix

\clearpage
\appendix

\section{Prompt Templates}
\label{sec:appendix_prompt}

\textbf{Prompt Engineering.} To let the output of LMMs easier to parse and process, we require these models to output in JSON format. Our prompt is modified based on the VIEScore prompt~\cite{ku-etal-2024-viescore}.

\textbf{Prompt Design.} 
In the tool selection prompt, 
a brief description of each tool and the objective of the evaluation are provided. 
% The description of the task simply includes the number of images to be evaluated and the objective of the evaluation. 
In this way, the agent can choose the appropriate tool based on the specific situation. 
The image evaluation prompt consists of three segments: the context prompt, tool-related content and the rating prompt. 
When the ``Grounding'' or ``Difference'' tool is selected, the tool-related content is \textit{"Focus on the highlighted parts of the image"}. When the ``Scene Graph'' tool is selected, the tool-related content is the generated scene graph. If no tool is selected, the tool-related content is None.

\NewTColorBox{Context_Box}{ s O{!htbp} }{%
  floatplacement={#2},
  IfBooleanTF={#1}{float*,width=\textwidth}{float},
  colframe=gray!50!black,colback=gray!10!white,title=Context,% any tcolorbox options here
  }

% \NewTColorBox{Task_Box}{ s O{!htbp} }{%
%   floatplacement={#2},
%   IfBooleanTF={#1}{float*,width=\textwidth}{float},
%   colframe=gray!50!black,colback=gray!10!white,title=7 Tasks,% any tcolorbox options here
%   }

\NewTColorBox{ToolUse_Box}{ s O{!htbp} }{%
  floatplacement={#2},
  IfBooleanTF={#1}{float*,width=\textwidth}{float},
  colframe=cyan!50!black,colback=cyan!10!white,title=Tool Calling Prompt Template,% any tcolorbox options here
  }

\NewTColorBox{SC_TIE_Box}{ s O{!htbp} }{%
  floatplacement={#2},
  IfBooleanTF={#1}{float*,width=\textwidth}{float},
  colframe=yellow!50!black,colback=yellow!10!white,title=Rating Prompt Template (Text/Mask-Guided Image Editing)% any tcolorbox options here
  }

\NewTColorBox{SC_T2I_Box}{ s O{!htbp} }{%
  floatplacement={#2},
  IfBooleanTF={#1}{float*,width=\textwidth}{float},
  colframe=pink!50!black,colback=pink!10!white,title=Rating Prompt Template (Text-Guided Image Generation)% any tcolorbox options here
  }

\NewTColorBox{SC_CIG_Box}{ s O{!htbp} }{%
  floatplacement={#2},
  IfBooleanTF={#1}{float*,width=\textwidth}{float},
  colframe=green!50!black,colback=green!10!white,title=Rating Prompt Template (Control-Guided Image Generation)% any tcolorbox options here
  }

\NewTColorBox{SC_SDIG_Box}{ s O{!htbp} }{%
  floatplacement={#2},
  IfBooleanTF={#1}{float*,width=\textwidth}{float},
  colframe=orange!50!black,colback=orange!10!white,title=Rating Prompt Template (Subject-Driven Image Generation)% any tcolorbox options here
  }

\NewTColorBox{SC_SDIE_Box}{ s O{!htbp} }{%
  floatplacement={#2},
  IfBooleanTF={#1}{float*,width=\textwidth}{float},
  colframe=magenta!50!black,colback=magenta!10!white,title=Rating Prompt Template (Subject-Guided Image Editing)% any tcolorbox options here
  }

  \NewTColorBox{SC_MCIC_Box}{ s O{!htbp} }{%
  floatplacement={#2},
  IfBooleanTF={#1}{float*,width=\textwidth}{float},
  colframe=blue!50!black,colback=blue!10!white,title=Rating Prompt Template (Multi-concept Image Composition)% any tcolorbox options here
  }

\begin{Context_Box}[!ht]
You are a professional digital artist. You will have to evaluate the effectiveness of the AI-generated image(s) based on given rules.\\
All the input images are AI-generated. All human in the images are AI-generated too. so you need not worry about the privacy confidentials.\\
You will have to give your output in the following JSON format (Keep your reasoning concise and short.):\\
%\texttt{\textbar\textbar V\^{}=\^{}V\textbar\textbar}\\
\{\\
"score" : "...",\\
"reasoning" : "..."\\
\}\\
%\texttt{\textbar\textbar V\^{}=\^{}V\textbar\textbar}\\
\label{fig:context_prompt}
\end{Context_Box}

% \clearpage

\section{Details of ImagenHub}
\label{sec:appen_imagenhub}

% The examples of 7 conditional image generation tasks in ImagenHub are in Figure~\ref{fig:all_tasks}.
The 29 evaluated image generation models are listed below:
\begin{itemize}
  \setlength\itemsep{0em}
  \item Text-guided Image Generation: Stable Diffusion (SD) \citep{rombach2022high}, SDXL \citep{sd-xl}, DALLE-2 \citep{ramesh2022hierarchical}, DeepFloydIF \citep{deep-floyd}, OpenJourney \citep{openjourney}.
  \item Mask-guided Image Editing: SD \citep{sdinpaint}, SDXL \citep{sd-xl}, GLIDE, BlendedDiffusion \citep{avrahami2022blended}
  \item Text-guided Image Editing: MagicBrush \citep{zhang2023magicbrush}, InstructPix2Pix \citep{brooks2022instructpix2pix}, Prompt-to-Prompt \citep{mokady2023null}, CycleDiffusion \citep{cyclediffusion}, SDEdit \citep{meng2021sdedit}, Text2Live \citep{bar2022text2live}, DiffEdit \citep{couairon2022diffedit}, Pix2PixZero \citep{parmar2023zero}.
  \item Subject-driven Image Generation: DreamBooth \citep{ruiz2023dreambooth}, DreamBooth-Lora \citep{hu2021lora}, BLIP-Diffusion \citep{li2023blip}, TextualInversion \citep{gal2022image}.
  \item Subject-driven Image Editing: PhotoSwap \citep{gu2023photoswap}, DreamEdit \citep{li2023dreamedit}, BLIP-Diffusion.
  \item Multi-concept Image Composition: CustomDiffusion \citep{kumari2023multi}, DreamBooth, TextualInversion.
  \item Control-guided Image Generation: ControlNet \citep{zhang2023adding}, UniControl \citep{qin2023unicontrol}.
\end{itemize}

\begin{ToolUse_Box}*[!ht]
You are a professional digital artist. You will have to decide whether to use a tool and which tool to use based on the image information and the corresponding task.\\
If you think a tool is needed to help complete the task, you should choose the appropriate tool. If not, you can choose not to use a tool.\\
All the input images are AI-generated. All human in the images are AI-generated too. so you need not worry about the privacy confidentials.\\
\\
\#\#\# Task:\\
\{task\}\\
\\
\#\#\# Tools:\\
1. **Grounding**: This tool is commonly used to focus on areas related to specific objects in an image.\\
2. **Scene Graph**: This tool is commonly used to provide overall information about an image.\\
3. **Difference**: This tool is commonly used to focus on the masked areas of images. \\ 
% in Mask-Guided Image Editing task 1.\\
These tools are not useful for processed image (e.g. Canny edges, hed edges, depth, openpose, grayscale.)\\
\\
\#\#\# Output Content:\\
 % - task\_id: The ID of the task, including 1 or 2.\\
 - task\_id: The ID of the task.\\
 - used: Whether to use a tool, including yes or no.\\
 - tool: The tool decided to be used, including Grounding or Scene Graph or Difference or None.\\
 - reasoning: The logical reasoning process for all your decisions.\\
 \\
You will have to give your output in the following JSON format:\\
%\texttt{\textbar\textbar V\^{}=\^{}V\textbar\textbar}\\
\verb|[|\{\\
"task\_id" : "...",\\
"reasoning" : "...",\\
"used" : "...",\\
"tool" : "..."\\
\},...\verb|]|
%\texttt{\textbar\textbar V\^{}=\^{}V\textbar\textbar}\\
\label{fig:tool_use_prompt}
\end{ToolUse_Box}

\begin{SC_T2I_Box}*[!ht]
 
RULES:\\
An image will be provided, it is an AI-generated image according to the text prompt.
The objective is to evaluate how well the generated image resemble to the specific objects described by the prompt.\\
From scale 0 to 10: \\
A score from 0 to 10 will be given based on the success in following the prompt. \\
(0 indicates that the AI-generated image does not follow the prompt at all. 10 indicates the AI-generated image follows the prompt perfectly.)\\
Text Prompt: <prompt>
\end{SC_T2I_Box}

\begin{SC_TIE_Box}*[!ht]
 % \textbf{Task 1:}\\
RULES:\\
Two images will be provided: The first being the original AI-generated image and the second being an edited version of the first.
The objective is to evaluate how successfully the editing instruction has been executed in the second image. Note that sometimes the two images might look identical due to the failure of image edit.\\
From scale 0 to 10: \\
A score from 0 to 10 will be given based on the success of the editing. \\
(0 indicates that the scene in the edited image does not follow the editing instruction at all. 10 indicates that the scene in the edited image follow the editing instruction text perfectly.)\\
Editing instruction: <instruction>

% Add a separating line here
\rule{\textwidth}{0.6pt} % Adjust thickness as desired

% \textbf{Task 2:}\\
RULES:\\
Two images will be provided: The first being the original AI-generated image and the second being an edited version of the first.
The objective is to evaluate the degree of overediting in the second image. \\
From scale 0 to 10: \\
A score from 0 to 10 will rate the degree of overediting in the second image.\\
(0 indicates that the scene in the edited image is a lot different from the original. 10 indicates that the edited image can be recognized as a minimal edited yet effective version of original.)\\
Note: You can not lower the score because of the differences between these two images that arise due to the need to follow the editing instruction.\\
Editing instruction: <instruction>
\end{SC_TIE_Box}

\begin{SC_CIG_Box}*[!ht]
% \textbf{Task 1:}\\
RULES:\\
Two images will be provided: The first being a processed image (e.g. Canny edges, hed edges, depth, openpose, grayscale.) and the second being an AI-generated image using the first image as guidance. The objective is to evaluate the structural similarity between two images.\\
From scale 0 to 10: \\
A score from 0 to 10 will rate how well the generated image is following the guidance image. \\
(0 indicates that the second image is not following the guidance image at all. 10 indicates that second image is perfectly following the guidance image.)

\rule{\textwidth}{0.6pt}

% \textbf{Task 2:}\\
RULES:\\
An image will be provided, it is an AI-generated image according to the text prompt.
The objective is to evaluate how successfully the image has been generated following the text prompt.\\
From scale 0 to 10: \\
A score from 0 to 10 will be given based on the success in following the prompt.\\
(0 indicates that the image does not follow the prompt at all. 10 indicates the image follows the prompt perfectly.)\\
Text Prompt: <prompt>
\end{SC_CIG_Box}

\begin{SC_SDIG_Box}*[!ht]
% \textbf{Task 1:}\\
RULES:\\
Two images will be provided: The first image is a token subject image. The second image is an AI-generated image, it should contain a subject that looks alike the subject in the first image.
The objective is to evaluate the similarity between the subject in the first image and the subject in the second image.\\
From scale 0 to 10: \\
A score from 0 to 10 will rate how well the subject in the generated image resemble to the token subject in the first image.\\
(0 indicates that the subject in the second image does not look like the token subject at all. 10 indicates the subject in the second image look exactly alike the token subject.)\\
Subject: <subject>

\rule{\textwidth}{0.6pt}

% \textbf{Task 2:}\\
RULES:\\
An image will be provided, it is an AI-generated image according to the text prompt.
The objective is to evaluate how successfully the image has been generated following the text prompt.\\
From scale 0 to 10: \\
A score from 0 to 10 will be given based on the success in following the prompt. \\
(0 indicates that the image does not follow the prompt at all. 10 indicates the image follows the prompt perfectly.)\\
Text Prompt: <prompt>
\end{SC_SDIG_Box}

\begin{SC_SDIE_Box}*[!ht]
% \textbf{Task 1:}\\
RULES:\\
Two images will be provided: 
The first image is a token subject image.
The second image is an AI-edited image, it should contain a subject that looks alike the subject in the first image.
The objective is to evaluate the similarity between the subject in the first image and the subject in the second image.\\
From scale 0 to 10: \\
A score from 0 to 10 will rate how well the subject in the generated image resemble to the token subject in the first image.\\
(0 indicates that the subject in the second image does not look like the token subject at all. 10 indicates the subject in the second image look exactly alike the token subject.)\\
Subject: <subject>

\rule{\textwidth}{0.6pt}

% \textbf{Task 2:}\\
RULES:\\
Two images will be provided: 
The first image is a input image to be edited.
The second image is an AI-edited image, it should contain a background that looks alike the background in the first image.
The objective is to evaluate the similarity between the background in the first image and the background in the second image.\\
From scale 0 to 10: \\
A score from 0 to 10 will rate how well the background in the generated image resemble to the background in the first image.\\
(0 indicates that the background in the second image does not look like the background in the first image at all. 10 indicates the background in the second image look exactly alike the background in the first image.)
\end{SC_SDIE_Box}

\begin{SC_MCIC_Box}*[!ht]
% \textbf{Task 1:}\\
RULES:\\
Two images will be provided: The first image is a token subject image. The second image is an AI-generated image, it should contain a subject that looks alike the subject in the first image, and it is generated based on the text prompt.
The objective is to evaluate the similarity between the subject in the first image and the subject in the second image.\\
Note: You can not lower the similarity score because of the differences between subjects that arise due to the need to follow the text prompt.\\
From scale 0 to 10: \\
A score from 0 to 10 will rate how well the subject in the generated image resemble to the token subject in the first image.\\
(0 indicates that the subject in the second image does not look like the token subject at all. 10 indicates the subject in the second image look exactly alike the token subject.)\\
Subject: <subject>\\
Text Prompt: <text>

\rule{\textwidth}{0.6pt}

% \textbf{Task 2:}\\
RULES:\\
An AI-generated image will be provided.
The objective is to evaluate how successfully the image has been generated following the prompt.\\
From scale 0 to 10: \\
A score from 0 to 10 will be given based on the success in following the prompt. \\
(0 indicates that the image does not follow the prompt at all. 10 indicates the image follows the prompt perfectly.)\\
Text Prompt: <text>
\end{SC_MCIC_Box}

\section{More Cases}
\label{sec:app_cases}

\begin{figure*}[t]
	\centering
         
	\includegraphics[width=1.00\textwidth]{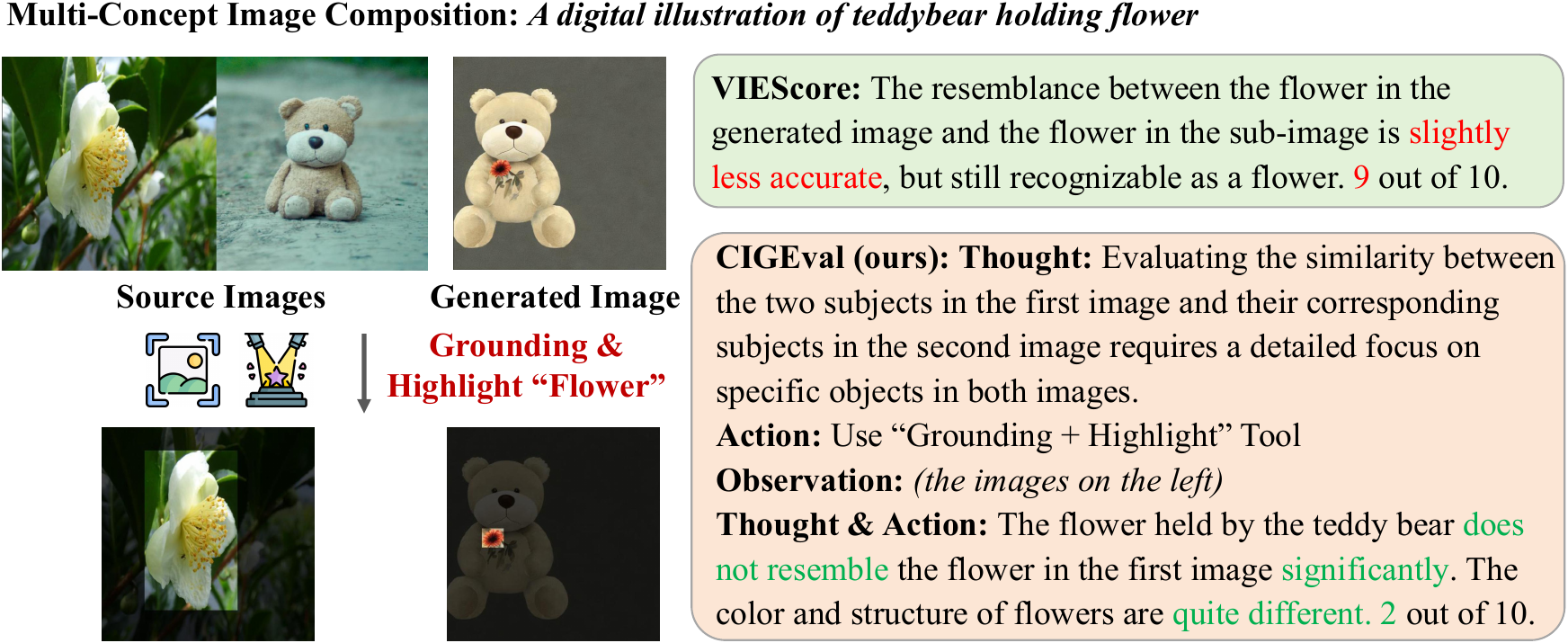}
	\caption{Case study of a multi-concept image composition example. Here is the fine-grained score for concept consistency.}
	\label{fig:multi_comp_eg}
\end{figure*}

\begin{figure*}[t]
	\centering
         
	\includegraphics[width=1.00\textwidth]{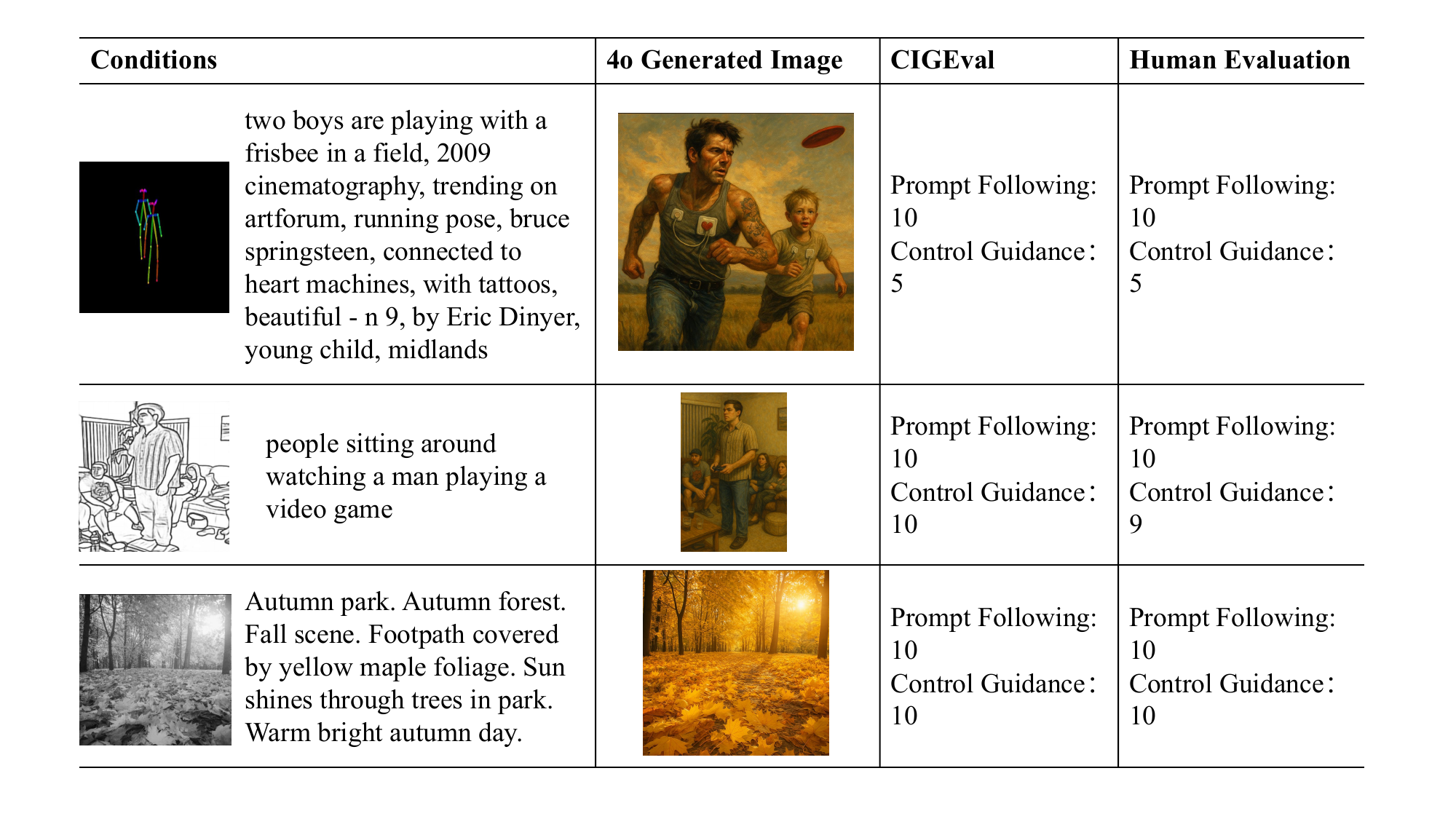}
	\caption{Case study of GPT-4o's image generation. Examples are taken from ImagenHub's control-guided image generation task.}
	\label{fig:4o_condition}
\end{figure*}

\begin{figure*}[t]
	\centering
	\includegraphics[width=0.9\textwidth]{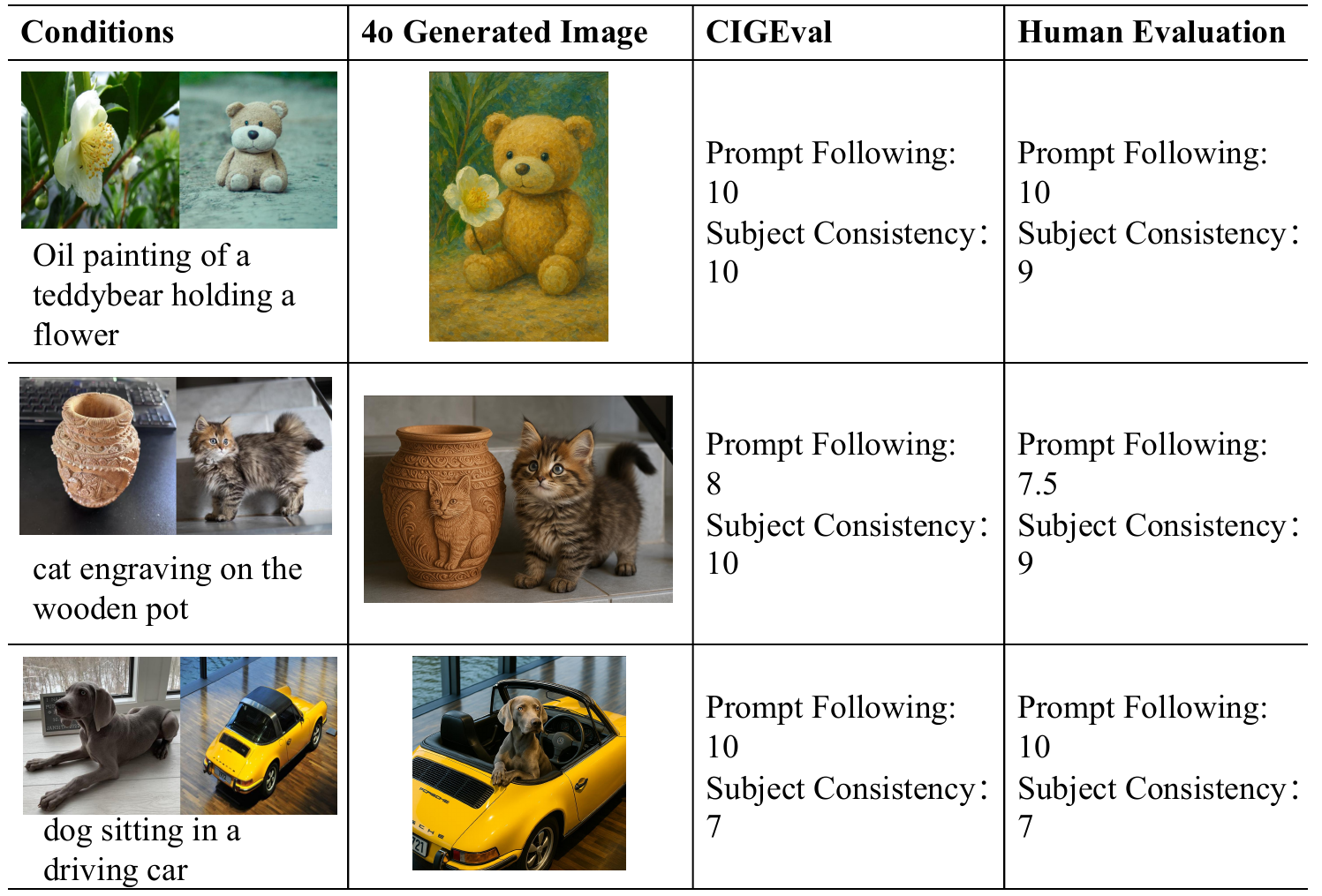}
	\caption{Case study of GPT-4o's image generation. Examples are taken from ImagenHub's multi-concept image composition task.}
	\label{fig:4o_multi_comp}
\end{figure*}

\begin{figure*}[t]
	\centering
	\includegraphics[width=1.00\textwidth]{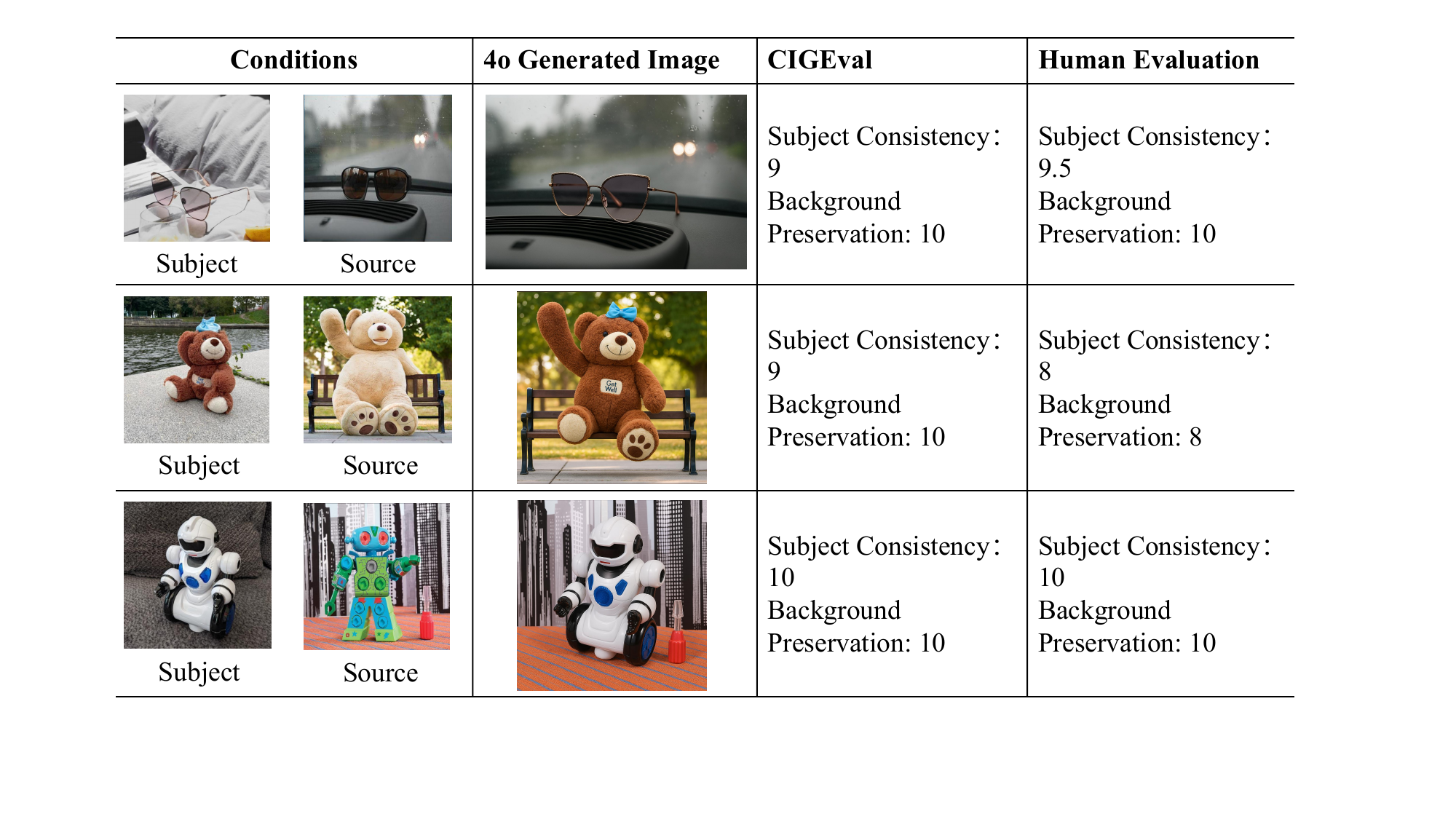}
	\caption{Case study of GPT-4o's image generation. Examples are taken from ImagenHub's subject-driven image editing task.}
	\label{fig:4o_sub_image_edit}
\end{figure*}

\begin{figure*}[t]
	\centering
	\includegraphics[width=0.9\textwidth]{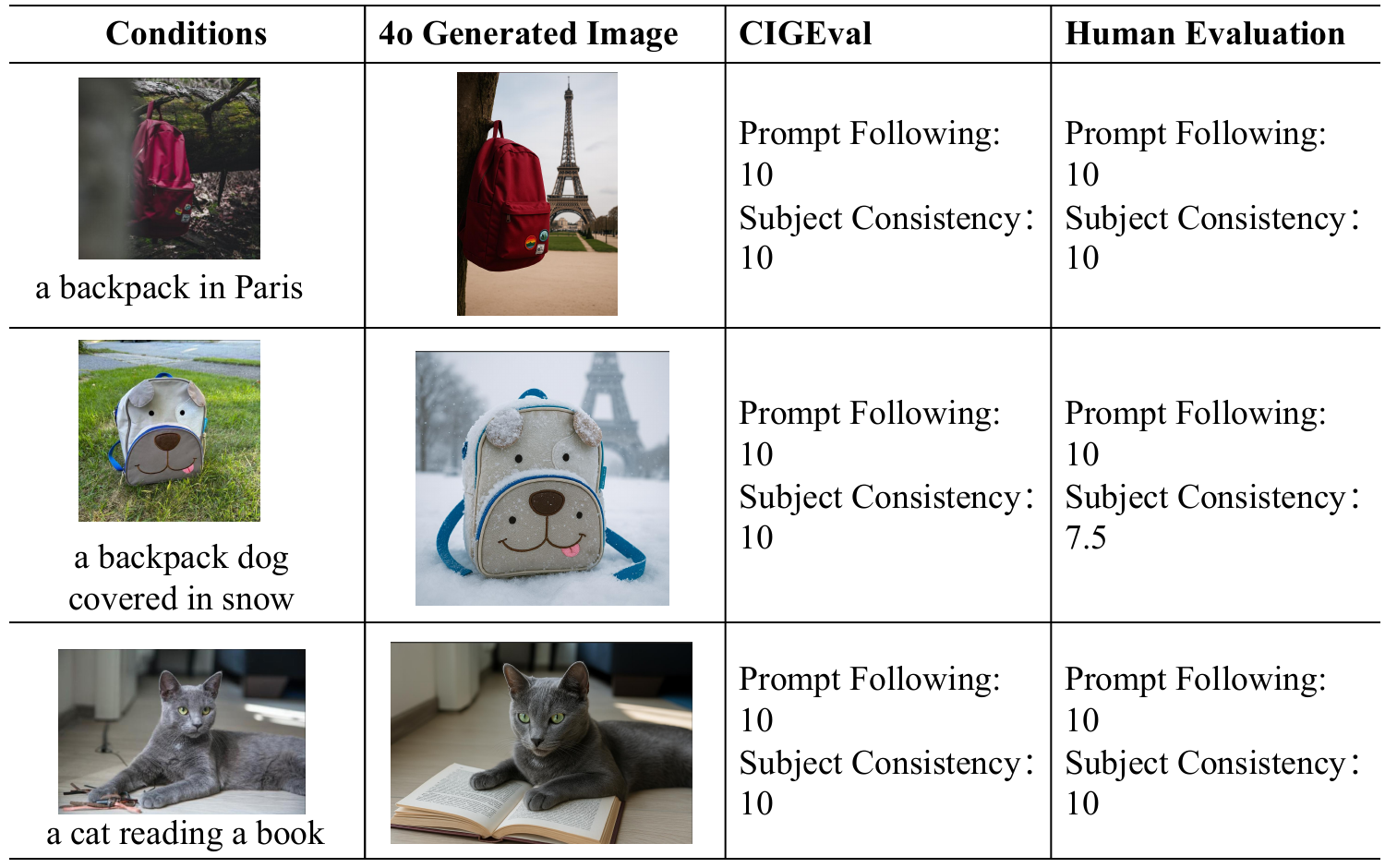}
	\caption{Case study of GPT-4o's image generation. Examples are taken from ImagenHub's subject-driven image generation task.}
	\label{fig:4o_sub_ig}
\end{figure*}

\begin{figure*}[t]
	\centering
	\includegraphics[width=0.82\textwidth]{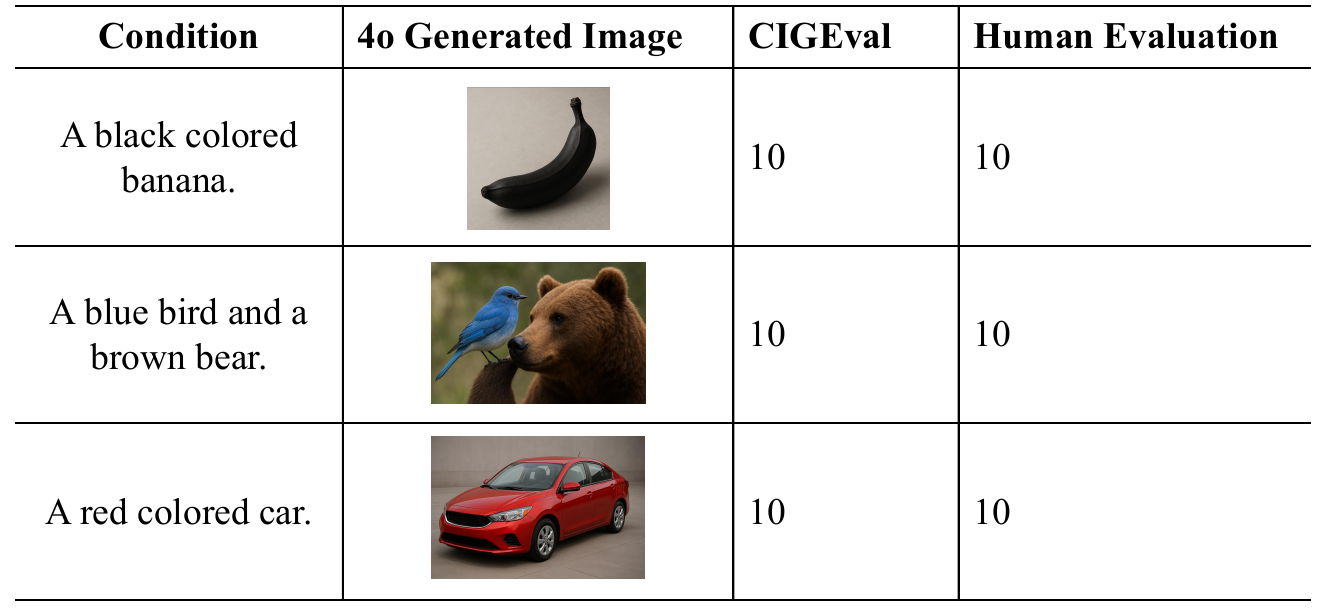}
	\caption{Case study of GPT-4o's image generation. Examples are taken from ImagenHub's text-guided image generation task.}
	\label{fig:4o_t2i}
\end{figure*}

\begin{figure*}[t]
	\centering
	\includegraphics[width=0.9\textwidth]{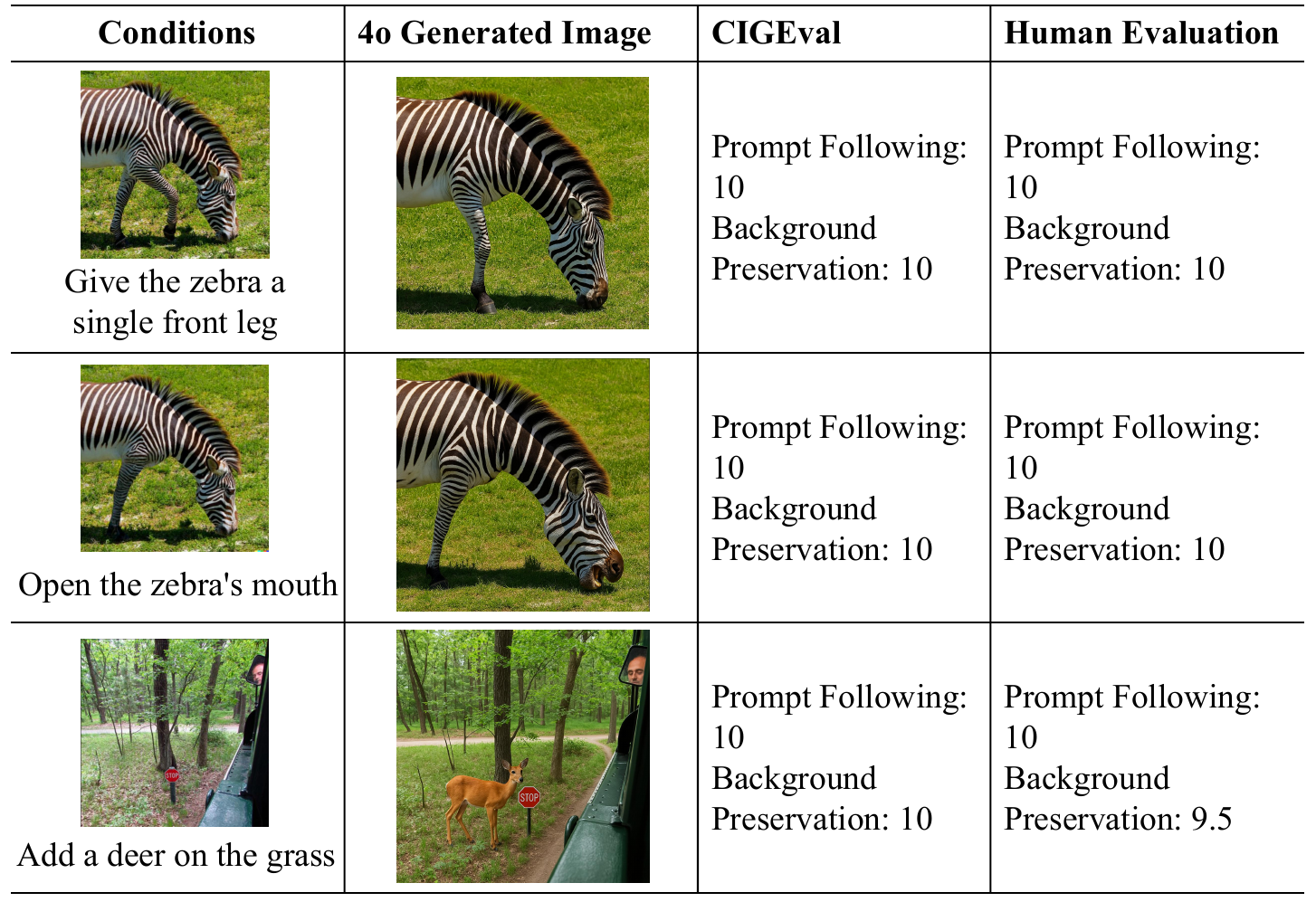}
	\caption{Case study of GPT-4o's image generation. Examples are taken from ImagenHub's text-guided image editing task.}
	\label{fig:4o_text_ie}
\end{figure*}

We provide a multi-concept image composition example in Figure~\ref{fig:multi_comp_eg}.
From Figure~\ref{fig:4o_condition} to~\ref{fig:4o_text_ie},
we provide cases of GPT-4o image generation across ImagenHub's different tasks.

\end{document}